# Landsat-Sentinel-2 algal bloom mapping using Vision Transformers: model description, implementation, and examples

Thainara M. A. Lima [a,*], Vitor S. Martins [a,*]

[a] Department of Agricultural & Biological Engineering, Mississippi State University (MSU), Starkville, MS 39762, USA.

*Corresponding author, E-mails: tml411@msstate.edu (Thainara M. A. Lima) and vmartins@abe.msstate.edu (Vitor S. Martins)

## Abstract

Coastal algal bloom monitoring requires frequent, spatially detailed, and globally consistent observations provided by Landsat-8/9 and Sentinel-2 A/B/C. Together, these missions provide more than a decade of medium-resolution multispectral imagery with near-global coverage every 2-3 days, enabling detection of fragmented bloom structures not resolvable by coarse ocean-color sensors. However, their use in aquatic environments remains challenging due to limited spectral coverage and the lack of harmonized reflectance products. As an alternative to traditional bio-optical methods, deep learning-based image classification provides a data-driven approach that can overcome many of these limitations. This study presents, for the first time, a successful implementation of vision transformer-based coastal algal bloom mapping using 30-m Landsat-Sentinel-2 images. A globally distributed algal bloom patch dataset (24,265 training and 12,255 validation patches, each 30m 256×256 pixels) was generated in coastal algal bloom-prone hotspots worldwide. Four transformer-based architectures, Vanilla Vision Transformer, Swin Transformer, SegFormer, and Masked Autoencoders, were implemented and compared against a standard convolutional baseline model (ResUNet) for fine-scale bloom detection. The models were assessed under different optical water types, atmospheric, and surface conditions. In general, all deep learning models demonstrated great capabilities in detecting floating algal bloom areas, with omission and commission errors ranging from 8% to 65%. When evaluated under cloud and glint stress conditions in a time series, the Swin Transformer outperformed traditional spectral-index approaches, which produced widespread false positives, while the deep learning model effectively avoided cloud- and glint-affected pixels. Additionally, comparisons with MODIS-derived products highlighted the benefits of a higher spatial resolution algal bloom product in detecting fragmented and irregularly affected bloom areas. Our findings support the new frontier of deep learning models as a reliable tool for medium-resolution, consistent monitoring of floating algal blooms in dynamic coastal environments.

## 1. Introduction

Algal blooms, defined as a rapid, large-scale accumulation of micro- or macroalgae in the upper water column, can occur across diverse coastal environments, with more than 5,000 species documented globally (Smith et al., 2003; Blondeau-Patissier et al., 2014; Gernez et al., 2023). Bloom events can emerge abruptly and are projected to increase with climate change, emphasizing the urgent need for a scalable coastal algal bloom monitoring (Anderson et al., 2009, 2012, 2019; Barton et al., 2016; Gobler, 2020; Hallegraeff et al., 2021). Satellite ocean color remote sensing provides multi-spectral datasets for detecting and tracking algal bloom events, documenting when and where they occur, and making it possible to understand and monitor their spatial and temporal dynamics (Gower et al., 2008; IOCCG, 2021; Dai et al., 2023). Ocean color sensors such as the Moderate Resolution Imaging Spectroradiometer (MODIS), Medium Resolution Imaging Spectrometer (MERIS), Medium Resolution Imaging Spectrometer (CZCS), Sea-viewing Wide Field-of-view Sensor (SeaWiFS), and Ocean and Land Color Instrument (OLCI) have demonstrated strong capabilities for monitoring high-biomass blooms at synoptic scales (Blondeau-Patissier et al., 2014; IOCCG, 2021). Additionally, several studies have been focusing on high-biomass blooms dominated by a single species in specific regions, for example, *Sargassum* in the Intra-Americas Sea and Atlantic (Gower and King, 2011; Gower et al., 2013; Wang and Hu, 2016), *Karenia brevis* in the Gulf of America (Tomlinson et al., 2009), and *Trichodesmium* in Australian waters (Blondeau-Patissier et al., 2018). Recently, Dai et al. (2023) demonstrated a global-scale bloom mapping using MODIS and unsupervised anomaly detection, which is one of the first studies providing daily bloom-affected area masks. While ocean color products support broad-scale monitoring, their moderate spatial resolution limits the detection of small, fragmented, or low-concentration blooms, which often form as clumps, mats, or rafts smaller than a single pixel (Ody et al., 2019; Qi and Hu, 2021; Wang et al., 2021).

Additionally, accurately retrieving water reflectance from coarse-resolution sensors remains challenging, particularly in coastal zones where bottom reflectance in shallow waters, high concentrations of suspended particles, and land adjacency effects introduce significant uncertainties (Odermatt et al., 2012; Bulgarelli and Zibordi, 2018; Wang and Hu, 2021a; Xu et al., 2023; Pahlevan et al., 2024). As a result, these uncertainties propagate into the problems of bloom detection using simple threshold-based spectral index mapping or deriving the chlorophyll-a concentrations (Darecki and Stramski, 2004; Hu, 2009; Blondeau-Patissier et al., 2014; Wang and Hu, 2021b, 2023; Podsosonnaya et al., 2025). As an alternative, attempts have been made to use high- and medium-resolution satellites, such as WorldView-2, Sentinel-2, Landsat-8/9, and Spot7 to study coastal areas at local scales (Qi et al., 2019; Qi and Hu, 2021; Wang and Hu, 2021b; Ma et al., 2021; Zhang et al., 2022). Medium spatial resolution sensors like NASA's Landsat-8/9 Operational Land Imager (OLI) and ESA's Sentinel-2 Multispectral Imager (MSI) offer global, consistent observations for scalable algal bloom products (Pahlevan et al., 2017a, 2019; Wang and Hu, 2021a). Their combined use through the Harmonized Landsat-Sentinel (HLS) virtual constellation enables near-global coverage every 2–3 days (Claverie et al., 2018; Ju et al., 2025; Zhou et al., 2025), bringing a unique opportunity for the timely detection of the rapid proliferation of algal blooms. Yet, the utility of HLS in aquatic environments demands specific image processing and radiometric corrections (Page et al., 2019; Cao et al., 2022), and implementing these methods is essential to expand the application of multi-mission satellite data for coastal monitoring. If successful, medium spatial resolution observations (5-30 meters) may enable mapping of small and irregular bloom areas, not detected by moderate (30-250 m) or coarse (>250 m) resolution sensors, revealing coastal bloom dynamics and seasonal patterns that were never explored (Hu et al., 2009; Hu et al., 2016; Cui et al., 2018; Wang and Hu, 2016; 2021a, 2021b; Descloitres et al., 2021; Qi and Hu, 2021; Zhang et al., 2022; Qi et al., 2023).

Through the years, several techniques have been developed to identify algal blooms using optical satellite observations, including band-ratio algorithms, spectral indices, bio-optical and analytical models, supervised classification, and principal component analysis (Kislik et al., 2022; Alharbi, 2023; Colkesen et al., 2024). Among these, spectral indices such as the Normalized Difference Chlorophyll Index (NDVI) (Mishra and Mishra, 2012; Kislik et al., 2022), Floating Algae Index (FAI) (Hu, 2009; Wang and Hu, 2016), Surface Algal Bloom Index (SABI) (Alawadi, 2010), and Maximum Chlorophyll Index (MCI) (Gower et al., 2007) are widely used to detect affected bloom pixels with empirical thresholds or to model chlorophyll-a concentrations (Tran et al., 2023). However, even minor uncertainties in retrieved water reflectance can significantly compromise the reliability of temporal analyses based on spectral indices (Morel and Belanger, 2006; Caballero, 2020; Zahir et al., 2024). As a result, these methods require extensive preprocessing to remove clouds and distinguish non-algal features with similar spectral signatures (Hu et al., 2023b). Additionally, while spectral indices only rely on spectral information, ignoring spatial context information, deep learning (DL) semantic segmentation algorithms have emerged as a powerful tool in remote sensing that can effectively use both spectral and spatial features for context-aware classification (Li et al., 2020; Wang and Hu, 2021b; Yao et al., 2024). These semantic segmentation models use a subset of images, i.e., image patches, as input data to explore hierarchical and discriminative features, such as edges, texture, and color, during classification (Lv et al., 2023). Although DL has proven applications in remote sensing (Ma et al., 2019; Arellano-Verdejo et al., 2019; Hordiiuk et al., 2019; Wang et al., 2019; Wang and Hu, 2021b; Cao et al., 2022; Laval et al., 2023), its use for algal bloom mapping remains emergent, with few studies detailing labeling protocols and no publicly available reference datasets.

State-of-the-art DL approaches predominantly rely on Convolutional Neural Networks (CNNs), which use learnable kernels to extract hierarchical spatial features from input images (Maggiori et al., 2017). Among these, the U-Net architecture, originally developed for biomedical segmentation, has proven effective for water body analysis due to its encoder-decoder structure and ability to capture multiscale contextual features (Jakovljevic et al., 2020; Qi et al., 2021; Wang and Hu, 2021b; Hu et al., 2023b; Yao et al., 2024; Wang et al., 2026). More recently, Vision Transformers (ViTs) have gained attention in remote sensing for their ability to model global relationships via self-attention mechanisms, surpassing the locality limitations of CNNs (Gao et al., 2021; Yang et al., 2022). To date, no studies have applied ViT models to algal bloom segmentation, and it remains unknown if they are useful to accurately detect bloom-affected areas and how well they can generalize across different optical water types; our study addresses these gaps. Similarly, foundation models have emerged as a new paradigm for geospatial AI (Lu et al., 2025). For instance, Prithvi is the latest IBM/NASA foundational model (Jakubik et al., 2023; Szwarcman et al., 2024) pre-trained on vast Earth observation datasets using self-supervised learning. These foundational models can be fine-tuned for a wide range of downstream tasks with minimum labeled data, offering scalability and transferability (Xiao et al., 2024). Its application was never evaluated in aquatic remote sensing, and exploring such a model in the context of algal bloom mapping holds a new frontier for coastal science and expands the possibilities with medium spatial resolution satellite data sources.

This study presents the first implementation of ViT models for 30-m coastal algal bloom mapping using harmonized Landsat-8/9 OLI and Sentinel-2 MSI data worldwide. The study has two primary objectives: *(i)* to integrate Landsat-Sentinel-2 images in a seamless water reflectance product and develop a globally representative 30-m algal bloom mask reference dataset, and *(ii)* implement and evaluate the applicability of novel vision transformers for 30-m coastal algal bloom mapping in distinct optically complex waters. We generated harmonized Landsat-8/9 and Sentinel-2 atmospherically corrected images (a total of 660 in 2023 and 186 in 2024) for aquatic applications, and created a globally distributed algal bloom patch dataset (24,265 patches for training and 12,255 for validation, with 30m 256×256-pixel patches) across diverse coastal environments on all continents. This algal bloom reference dataset was then used as the foundation for training and benchmarking a suite of state-of-the-art DL models, including Transformer-based architectures (e.g., Vanilla ViT, Swin Transformer, and SegFormer), Foundational Model Prithvi, and encoder-decoder CNNs (e.g., ResUNet). We evaluated the models' performance and generalization under different scenarios, including bloom-affected areas near high suspended solid plumes, clouds, and under sunglint effects. We compared our results with traditional threshold-based spectral indexes, and a moderate spatial resolution algal bloom product. Note that bloom events are highly heterogeneous in space and time, from greenish to reddish colors, depending on the dominant species (e.g., diatoms or dinoflagellates) and the concentration of water constituents (Dierssen et al., 2006). Kutser (2009) emphasizes the relative nature of the term "bloom," which can refer to phytoplankton or macroalgal events with varying biomass levels. This study focuses on algal blooms (microalgae or macroalgae) that produce saturated, spatially distinct, and textured patterns on the water surface, aiming to develop a generic, scalable detection approach, and we do not intend to characterize phytoplankton groups, community composition, or relative abundance. The main contributions of our study are: (i) the first application of Vision Transformer architectures for 30-m Landsat–Sentinel algal bloom mapping, demonstrating, for the first time, their capability to delineate bloom features in coastal waters; (ii) the development of a globally distributed 30-m algal bloom mask dataset, together with a scalable semi-automatic labeling workflow capable of producing consistent reference masks; and (iii) represents the first step toward a potential operational 30-m algal bloom mapping (geographic location, shape, impacted area) in coastal waters, which has never been performed.

## 2. Landsat-8/9 OLI and Sentinel-2 MSI data

### 2.1. Data description

In this study, we integrated multispectral observations from Landsat-8/9 OLI and Sentinel-2 MSI for 30-m coastal algal bloom mapping. The NASA Landsat program has provided the longest continuous record of Earth observations from space, spanning over 50 years since its launch in 1972 (Wulder et al., 2022). Landsat-8 and Landsat-9 were launched in 2013 and 2021, respectively, and operate in a sun-synchronous orbit with a 16-day revisit cycle each (Roy et al., 2014; Masek et al., 2020; Kabir et al., 2023). Both satellite platforms carry the Operational Land Imager (OLI), a multispectral pushbroom sensor with 9 spectral bands across the visible, NIR, and SWIR wavelengths at 30-meter spatial and 12-bit radiometric resolutions (Loveland & Irons, 2016). Together, the two satellites offer an 8-day revisit frequency, using the World Reference System-2 and Universal Transverse Mercator (UTM) projection. Since 2016, data from Landsat-8/9 have been distributed as Collection 2 (Claverie et al., 2018), with Level-1 products providing calibrated digital numbers in 185×185 km tiles that can be converted to top-of-atmosphere (TOA) reflectance.

The Sentinel-2 mission, developed by the European Space Agency (ESA) under the Copernicus program, comprises three satellites: Sentinel-2A (launched in 2015), Sentinel-2B (2017), and the more recent Sentinel-2C (2024) (Drusch et al., 2012). Each satellite carries the Multispectral Instrument (MSI) sensor, which captures imagery across 13 spectral bands and delivers products in UTM-based Military Grid Reference System (MGRS) tiles at various processing levels. MSI offers variable spatial resolution: 10 m for four VNIR bands, 20 m for six Red Edge, narrow NIR, and SWIR bands, and 60 m for three atmospheric correction bands. Designed for medium spatial resolution optical land monitoring, the mission provides a 5-day revisit interval and a 290 km swath width (Drusch et al., 2012). The primary product, Level-1C Collection 1, contains top-of-atmosphere (TOA) reflectance data in 109 × 109 km tiles. Combined, Landsat-8/9 and Sentinel-2A/B offer synergistic capabilities and a global median revisit time of 2-3 days (Li and Roy, 2017; Zhou et al., 2025).

## 2.2. Coastal algal bloom hotspot selection and data acquisition

For training algal bloom dataset generation (Section 3.3), 225 high-algal-occurrence regions were identified across coastal zones, which extend up to 370 km from coastlines, based on a global daily MODIS-based algal bloom occurrence product developed by Dai et al. (2023). First, we derived a long-term 2003-2020 frequency composite map of hotspots to identify candidate regions for algal bloom occurrences, highlighted as reddish areas in Figure 1A, ensuring a worldwide, geographically diverse search under varying environmental conditions. All available 2023 Sentinel-2 A/B and Landsat-8/9 images with less than 30% cloud cover were acquired from Google Earth Engine (GEE) for the selected locations. This initial search resulted in a total of 15,421 images, from which the Normalized Difference Vegetation Index (NDVI) ($\mathrm{NDVI} = (\mathrm{NIR} - \mathrm{Red})/(\mathrm{NIR} + \mathrm{Red})$), FAI ($\mathrm{FAI} = \mathrm{Red} + (\mathrm{SWIR} - \mathrm{Red}) \times (\lambda_{\mathrm{NIR}} - \lambda_{\mathrm{Red}})/(\lambda_{\mathrm{SWIR}} - \lambda_{\mathrm{Red}})$), RGB composites (red, green, blue), and false color composites (red, NIR, green) were generated for each image. A visual inspection was then conducted, ensuring the inclusion of both micro- and macroalgal occurrences, from which 660 images with visible algal bloom were selected. For our experiments, a single recent year (2023) provided a sufficiently large number of algal bloom events for a representative training dataset, encompassing a complete annual cycle with images in all seasons, under a broad range of climatic conditions, and varying sun and sensor angles, which enables the model to learn robust spectral and spatial representations of algal bloom events worldwide. Furthermore, by using a rigorous selection in hotspot regions, we minimized the risk of underrepresenting blooms that may vary in their characteristics from year to year, since these locations are consistently prone to algal bloom events. Examples of algal bloom occurrences are highlighted in Figure 1C. Figure 1A and 1B show the location of each selected image and the number of observations per location. We downloaded a total of 281 Level-1 TOA images from Landsat (146 from Landsat-8, and 135 from Landsat-9) via the USGS EarthExplorer platform using the USGS/EROS API, and 379 Level-1C TOA reflectance images from Sentinel-2 (184 from Sentinel-2A, and 195 from Sentinel-2B), through the Copernicus Open Access Hub API. These 660 images (85.9 GB) from 2023 were used in the training dataset generation.

For validation algal bloom dataset generation, images were selected from 2024 to evaluate the temporal transferability of the model in a year not seen during training. Additionally, optical water types (OWTs) were included as an additional criterion for selecting validation sites to assess the model's generalization and applicability across diverse water conditions. We used the 21 OWTs spectral signatures defined by Spyrakos et al. (2018) and grouped them into three broader categories using k-means clustering. The categories were defined as: (i) phytoplankton-rich waters with elevated chlorophyll concentrations (OWTs 1, 2, 6, 7, 8, 10 and 11 in the original paper), (ii) turbid waters dominated by high suspended sediment loads (OWTs 4, 5, 9, 12), and (iii) optically clear waters with low levels of both particulate and dissolved organic matter (OWTs 3 and 13). For each class, the mean reflectance spectrum was simulated to MODIS bands and used as references in the Spectral Angle Mapper (SAM) classifier (Kruse et al., 1993). The classifier was applied to the mean annual composite of 1 km MODIS (MYDOCGA) 16-day $R_{rs}$ data available on GEE to generate a global 2024 OWT map. This map was then used to guide the selection of sites representing different water types. A total of 77 OLI images (33 from Landsat-8 and 44 from Landsat-9) and 109 MSI images (49 from Sentinel-2A, 58 from Sentinel-2B, and 2 from Sentinel-2C) were selected and acquired in 2024. These 186 images (40.5 GB) were used in the validation dataset generation. By focusing on recent image dates (2023 & 2024, Figure 1B), our training and validation datasets reflect the most recent algal bloom events and their characteristics. Figure S1A presents the mean $R_{rs}$ spectra of each selected image classified according to the three defined OWTs using SAM, and Figure 1A shows the geographic distribution of the validation images.

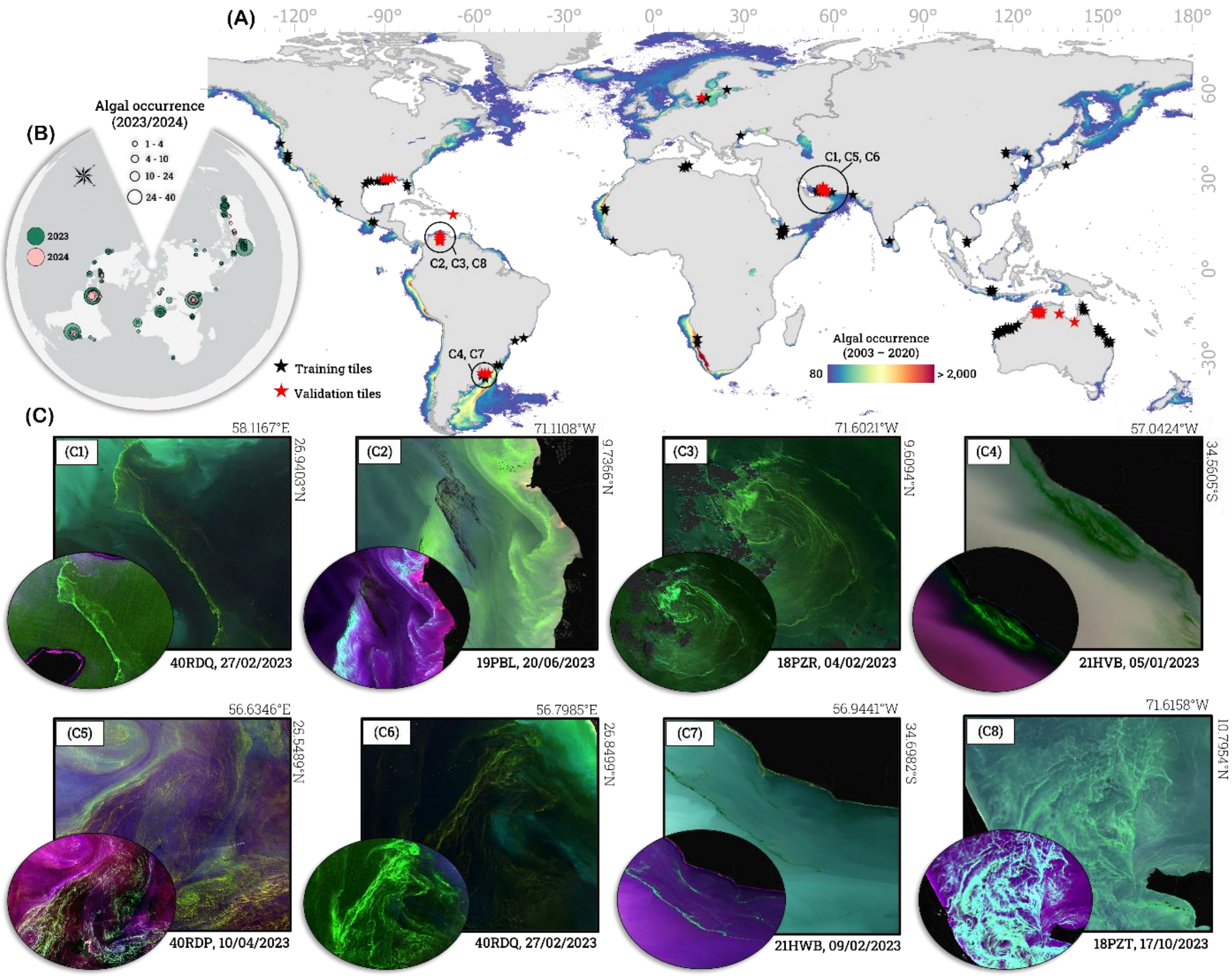

Figure 1. Coastal algal bloom hotspots for Landsat and Sentinel data selection, and examples of algal bloom events. (A) Sentinel-2/MSI tiles selected for algal bloom reference data generation for training (black stars) and validation (red stars). The colored map in the background shows bloom occurrence frequency between 2003-2020, generated from Dai et al. (2023). (B) Number of algal bloom occurrences identified in 2023 (green) and 2024 (rose golden), per tile, during visual inspection. (C) Examples of visually identified blooms. Each example shows the centroid coordinates (corners), MGRS tile ID, and acquisition date (bottom), and bloom location as indicated in (A). For every true-color image, a corresponding false-color composite (RED, NIR, GREEN) is displayed in the circle.

## 3. Methodology

### *3.1. Framework overview*

The workflow is illustrated in Figure 2. First, raw OLI and MSI TOA reflectance data were processed by applying our atmospheric correction pipeline, called AQUAVis, to retrieve remote sensing reflectance ($R_{rs}$), defined as the ratio of water radiance to the total downwelling irradiance just above the water surface. In the visible spectrum, $R_{rs}$ is shaped by the optically active constituents (OACs), providing a fundamental measure of water optical properties and enabling consistent comparison across sensors (Mobley, 1999). The retrieved $R_{rs}$ imagery was then used to harmonize OLI and MSI data into a single virtual constellation product for aquatic applications. Second, AQUAVis-derived $R_{rs}$ serve as input data to a semi-automatic algal bloom labeling procedure, where a dynamic NDVI-based threshold and an unsupervised object detection approach were combined to generate binary ground truth masks of algal blooms. In the third step, $R_{rs}$ images and labeled masks were tiled into patches and used to train and validate different deep learning architectures. This phase included quantitative accuracy assessment and area-based comparisons of model outputs. Finally, the best-performing architecture was applied to produce a globally applicable algal bloom segmentation model, which was further evaluated under diverse atmospheric and water optical conditions.

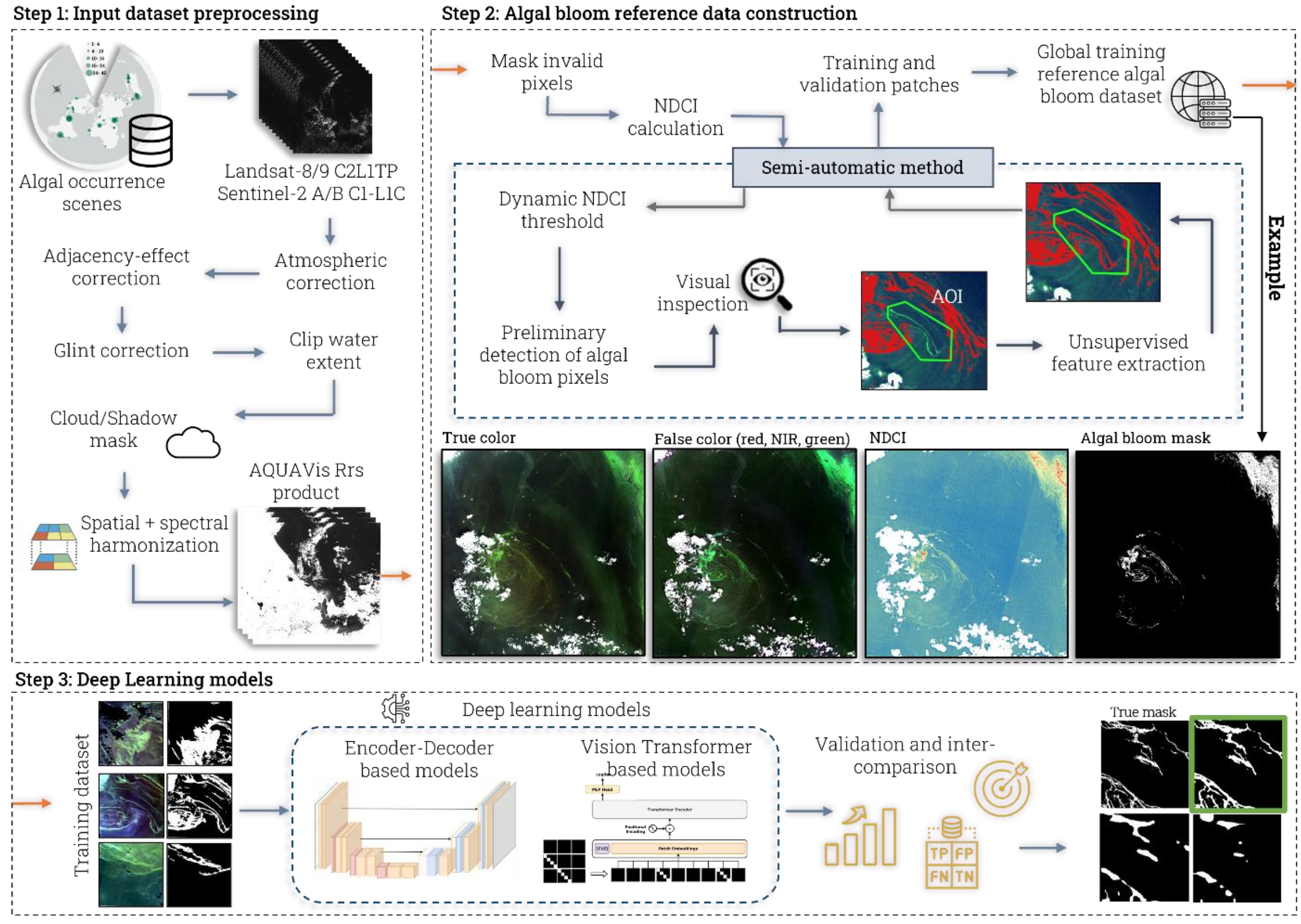

Figure 2. General workflow

### *3.2. Input dataset pre-processing*

We generated a harmonized Landsat-Sentinel-2 $R_{rs}$ data following the pipeline proposed by Lima et al. (2025b), called AQUAVis (Aquatic Virtual Constellation for Landsat-Sentinel $R_{rs}$ Observations). Following NASA's Harmonized Landsat-Sentinel product concept, AQUAVis is a globally validated processing framework designed to produce a harmonized Landsat 8/9 and Sentinel-2A/B reflectance dataset for aquatic remote sensing applications (Lima et al., 2025b). The pipeline integrates atmospheric correction using the 6SV (Second Simulation of a Satellite Signal in the Solar Spectrum Vector) (Vermote, 1997a) following the approach proposed by Gordon and Wang (1994). Additional corrections were applied to mitigate glint and adjacency effects (Tanré et al., 1981; Vermote et al., 1997b). The framework also applies sensor-specific algorithms for OLI and MSI data, including reprojection, spatial convolution resampling, gridding, and spectral bandpass adjustments derived from a globally representative spectral database of over 4,000 water bodies (Lima et al., 2025a). Additionally, the pipeline generates a quality assurance (QA) band that includes cloud and shadow masks derived using the FMask algorithm (Zhu et al., 2015). The QA band also incorporates a water mask, created by thresholding a water-related spectral index in combination with the SWIR band. Details of each step are described by Lima et al. (2025b). The result is a harmonized Landsat-Sentinel-2 $R_{rs}$ in four spectral bands (blue 483 nm, green 560 nm, red 660 nm, and NIR 865 nm), delivered at 30-meter spatial resolution in the UTM projection using Military Grid Reference System (MGRS). In Lima et al. (2025b) AQUAVis $R_{rs}$ product is validated using 2,664 AERONET-OC matchup observations across 23 global sites, showing spectral differences ranging from 0.39% in the blue to 21% in the NIR, with median biases near zero ($\leq$0.0014 $sr^{-1}$) and interquartile ranges within ±0.002 $sr^{-1}$, indicating excellent radiometric consistency. Full radiometric assessment of the AQUAVis-derived $R_{rs}$ is beyond the scope of this study, and all details regarding its processing, accuracy, and reliability are documented in Lima et al. (2025b). While the original AQUAVis pipeline did not include SWIR bands in the harmonization process, we derived atmospherically corrected SWIR bands and incorporated them into the pipeline given its critical role in spectral index calculations. A total of 660 (from 2023 for training) and 186 (from 2024 for validation) $R_{rs}$ images were generated to deliver the algal bloom patch training and validation datasets. The AQUAVis pipeline for the 847 images ran in the Mississippi Center for Supercomputing Research at the University of Mississippi using 50 x 2-Core Intel E5 2650.

### *3.3. Generation of Landsat-Sentinel-2 algal bloom patch reference datasets for training and validation*

We developed a semi-automatic algal bloom patch labeling approach to generate the first medium spatial resolution algal bloom patch reference datasets, where 30-meter pixels were labeled as either (1) algal bloom or (0) background. The first step was masking invalid pixels, such as land and pixels affected by clouds, shadows, areas adjacent to clouds, and strong sunglint (Hu et al., 2023b). Cloud, shadow, and land pixels were masked using the QA band generated by the AQUAVis pipeline (Section 3.2), with an additional 30-meter buffer applied to mask adjacent pixels. To minimize nearshore background interference, a 200-meter buffer was applied from the shoreline into the water. Since AQUAVis water masks were generated using a SWIR-based threshold and included glint correction, no additional glint mask was applied. NDVI was computed as (R865 − R665)/(R865 + R665) to identify algal bloom candidates (Yao et al., 2024), as the index has been successfully applied for algal bloom detection (Kahru et al., 1993; Ma et al., 2007; Oyama et al., 2015; Qin et al., 2022; Hu et al., 2023a). To determine the index's sensitivity to water reflectance changes and to identify the optimal threshold, we applied a series of NDVI values, ranging from -0.6 to 0.3 in 0.05 increments. The resulting masks were visually inspected to identify the threshold that best captured the observable bloom regions. To prevent the propagation of artifacts, we followed standard DL calibration practices by manually refining the initial labels (Wang & Hu, 2021; Hu et al., 2023). This step decouples the ground truth from the spectral index magnitude alone, ensuring that the labels remain independent of spectral index instability (Yao et al., 2024; Qi et al., 2025). Specifically, masks were refined using false-color composites (Red, NIR, Green), where floating algae are clearly distinguished by their high NIR reflectance (Qi et al., 2020). Bloom-like features were identified based on morphological cues, such as wispy, filamentous structures extending downstream. Not surprisingly, this is a complex task because algal blooms often exhibit a diffuse nature (i.e., dispersed unevenly over large areas) and lack clearly defined boundaries due to gradual transitions and patterns across the water surface. To address this, misclassified regions were manually defined via coarse polygons, as illustrated by the green polygon in Figure 2, followed by the application of an unsupervised object detection method (Wang and Hu, 2015). This method involved Gaussian filtering to suppress noise, followed by histogram equalization and a standard deviation-based threshold to enhance and segment the bloom region. The corrected pixels were then integrated into the original bloom mask. A final round of visual inspection ensured the reliability and consistency of the labeled algal bloom areas. Figure 2 shows an example of the final mask alongside true-color, false-color, and original NDVI imagery.

A total of 24,265 patches, each with 30m 256 × 256 pixels, were extracted from 2023 images for the algal bloom training dataset. Following the methodology proposed by Qi et al. (2025) and Yao et al. (2024), the six input band variables are blue, green, red, and near-infrared (NIR) bands from the Landsat-Sentinel $R_{rs}$ product, along with NDVI and FAI. We conducted a sensitivity test (Appendix S2) to determine if omitting FAI and NDVI from the model inputs would yield comparable performance. As shown in Figure S10, NDVI and FAI support the distinction between floating algal blooms and water, and lead to better edge detection of algal blooms (Sun et al., 2024). Within the deep learning framework, the spectral indexes are included as spatial contrast channels, i.e., two additional dimensions of the token embedding that enhance the spectral separability between bloom and water. Rather than providing absolute physical quantities, these indices contribute a sharpened edge-detection feature, and incorporating these indices as additional bands improves segmentation performance. Figure 3 illustrates representative examples of the extracted training patches. The number of patches per image was determined by multiplying the ratio of surface water to patch area by a factor of 1.5, ensuring comprehensive coverage of bloom structures. Patch centroids were randomly sampled from water pixels with a minimum 128-pixel separation. This deliberate overlap provides implicit translational augmentation, forcing the model to learn position-independent features rather than location-specific artifacts (Martins et al., 2022). Each candidate patch was defined as a 256 × 256-pixel box centered on a generated centroid and retained only if it included at least 10 algal bloom pixels. The training dataset was divided into 85% training (20,572) and 15% validation (3,693) subsets; note that this small validation set was only used to measure the training progress. The same methodology of patch extraction was applied to derive the validation dataset, but we used the independent 186 images from 2024, resulting in 12,255 validation patches. Histograms of bloom area proportions (Figure S2) demonstrate that the training and validation datasets are independently and representatively partitioned across all bloom sizes.

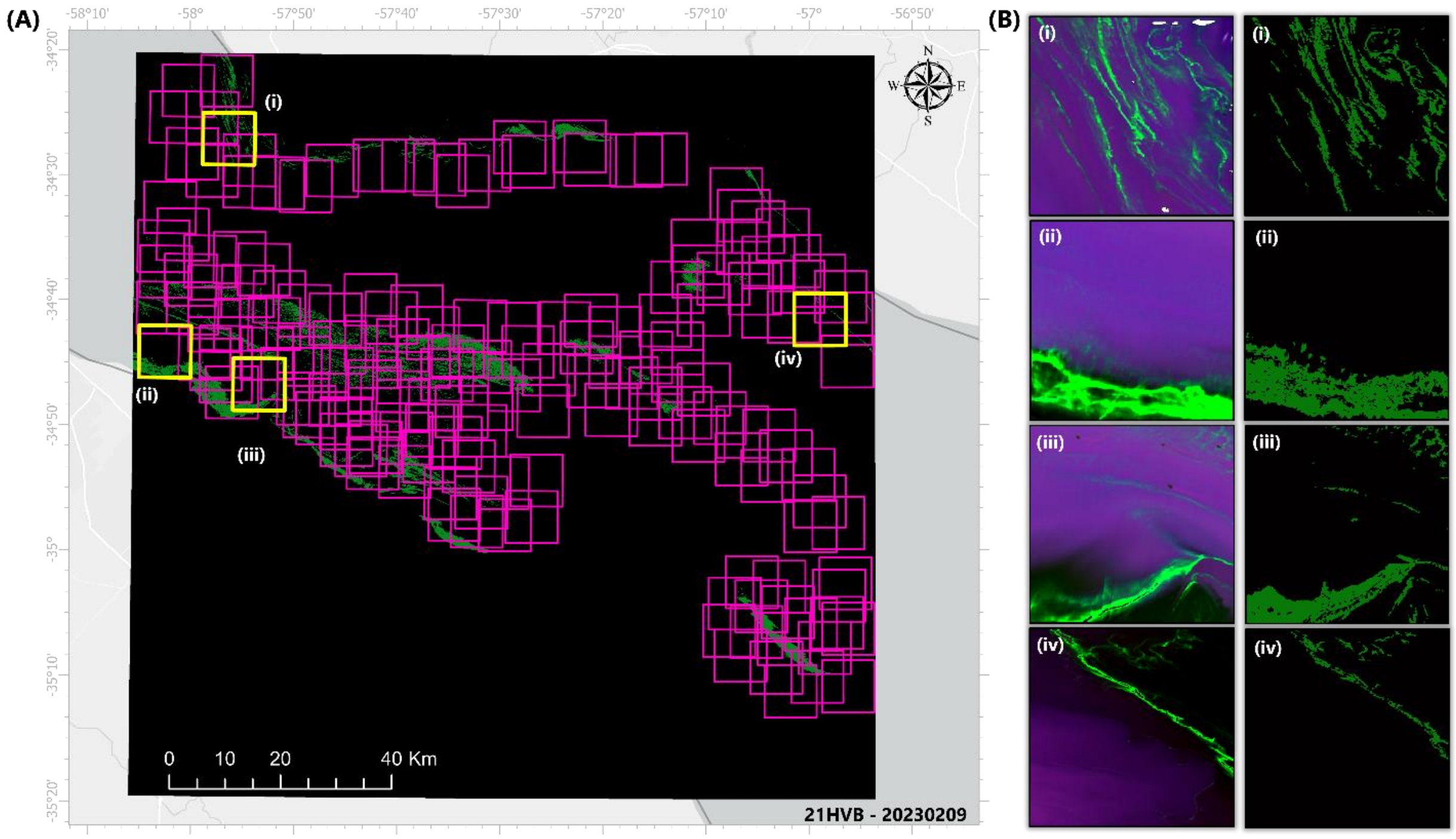

Figure 3. Illustration of the patch extraction process used to construct the training and validation dataset. (A) shows 256×256-pixel patches (magenta bounding boxes) sampled from the input ground-truth algal bloom mask (in green), following the spatial sampling strategy described in Section 3.3. (B) highlight examples of extracted patches, with each pair consisting of a false-color composition (red, NIR, green) and its corresponding binary algal mask. Ground truth algal bloom mask is located in MGRS tile ID 21HVB (57.53°W, 34.78°S) on February 09, 2023.

### *3.4. Deep learning algorithms training for algal bloom pixel extraction*

We implemented and evaluated five deep learning models for image segmentation using CNN-based and transformer-based architectures. Given the novelty of these models for the aquatic community, we provided a detailed description of their concept, structure, and configurations. All models were trained with a 24,265 algal bloom patch reference dataset using an NVIDIA GPU A100 80GB card.

#### *3.4.1. ResUNet architecture with context encoding*

The U-Net architecture (Ronneberger et al., 2015) is the most widely adopted CNN-based model for segmentation in remote sensing tasks (Kim et al., 2019; Gao et al., 2021; Hu et al., 2023b; Wang and Hu, 2021b; Feng et al., 2022; Yao et al., 2024). Based on the model developed by Diakogiannis et al. (2020), we proposed here the ResUNet with context encoding (Figure 4). Its structure comprises an encoder for hierarchical feature extraction (Figure 4, left side), a bottleneck for high-level semantic compression, and a decoder for spatial reconstruction (Figure 4, right side). The architecture is also enhanced by skip connections to fuse multi-scale features and residual layers to mitigate vanishing/exploding gradients (Xiao et al., 2018). In the bottleneck (Figure 4, bottom), we incorporated dilated convolutions to capture wide-context features (Yu et al., 2015) and the Context Encoding Module to apply channel attention (Zhang et al., 2018). The final decoder block produces a 256×256 feature map used for algal bloom prediction. Details on model structure are described by Martins et al. (2022).

During model training, key hyperparameters were tuned using Bayesian optimization (Brochu et al., 2010), resulting in a learning rate of 5e-5 and weight decay of 2e-4. AdamW optimizer was used with a cosine scheduler to gradually reduce the learning rate during training. Conventional image normalization was applied to normalize the input features using the global minimum and maximum, calculated per band. Basic data augmentations, including random rotations and flips, helped improve robustness. An early stopping criterion based on validation loss was implemented to avoid overfitting (Liu et al., 2008). The model was trained using a composite loss function combining Dice loss, Cross Entropy Loss, and IoU loss to address class imbalance and improve boundary precision. Additionally, a distance-based weighting map derived from the ground truth mask was incorporated into the loss to emphasize object borders (Ronneberger et al., 2015).

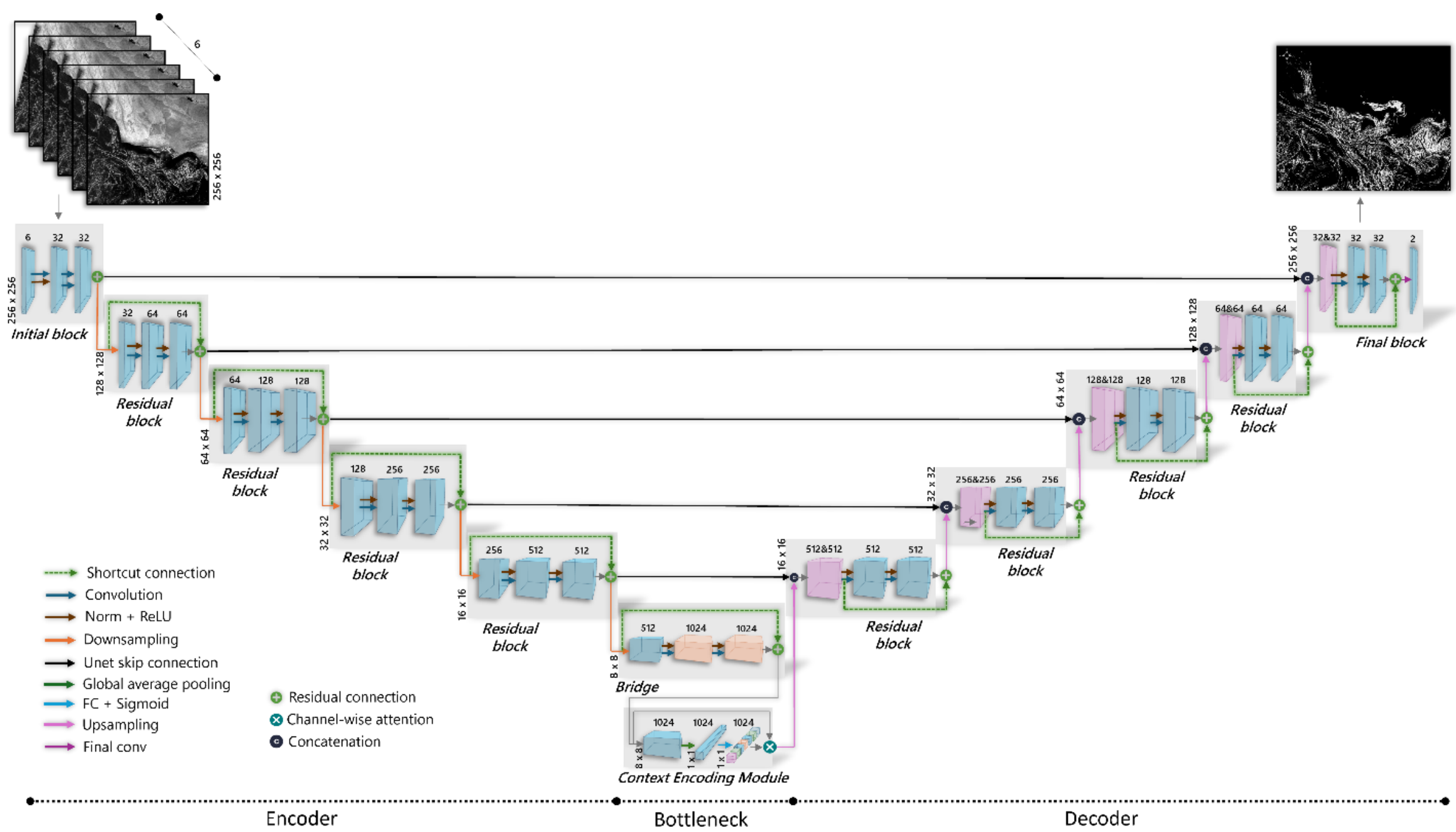

Figure 4. ResUNet architecture with context encoding used to segment 256 × 256 pixel patches for algal bloom mapping. The model employs 3×3 convolutional kernels and five encoder–decoder blocks that progressively increase the number of filters (up to 1024) while reducing spatial resolution. Residual connections (green arrows) facilitate gradient flow and stabilize training, while the decoder progressively upsamples and fuses encoder features to reconstruct spatial details. The number of feature maps is indicated above each box, and spatial dimensions are shown to the left of each block.

### *3.4.2. Pre-trained Vanilla Vision Transformer*

Vision Transformers (ViTs) have recently emerged as a powerful alternative to CNNs for dense prediction tasks such as semantic segmentation. In a ViT framework (Figure 5), the input image is divided into overlapping sub-patches, which are flattened and linearly projected into an embedding space, forming a sequence of tokens. These tokens are passed through stacked transformer blocks, where self-attention captures long-range contextual dependencies and builds global feature representations (Dosovitskiy et al., 2021; Vaswani et al., 2017). For semantic segmentation, a decoder and segmentation head reconstruct the spatial structure to generate a pixel-wise classification map (Zheng et al., 2021). For ViT model training, the inputs were those labeled algal bloom features, together with the four $R_{rs}$ AQUAVis bands, NDVI, and FAI. Considering the size of our dataset, we initialized the model with pretrained ImageNet-1k weights (Russakovsky et al., 2015) to enhance the model's ability to capture spatial structures while reducing training demands (Rombado et al., 2024). Details on our architecture structure are provided in the supplemental material (Appendix S1). Training was performed using AdamW optimizer with a learning rate and weight decay of 1e-4, guided by a cosine scheduler. Model optimization, input normalization, and augmentations followed the approach in Section 3.4.1. To improve training efficiency and stability using the pretrained weights, we first froze the encoder to train only the decoder, then unfroze it after several epochs with a lower learning rate to preserve pretrained features while fine-tuning alongside the decoder (Lee et al., 2019). This gradual fine-tuning ensured stable convergence, as the pretrained weights are adjusted incrementally to the target task while maintaining their generalization capabilities (Lima and Marfurt, 2020).

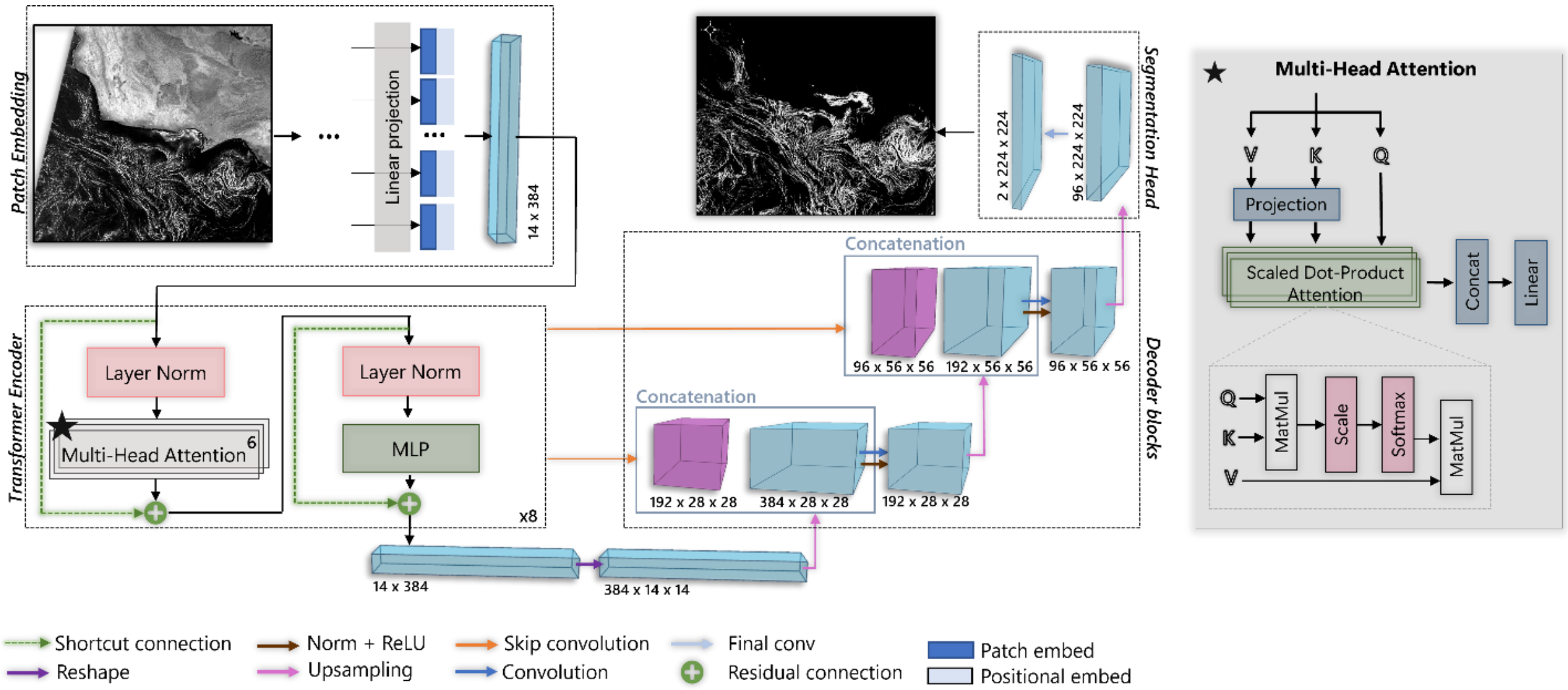

Figure 5. ViT architecture adapted for the algal bloom task. The input patch is split into smaller non-overlapping spectral sub-patches and linearly projected into embeddings. These embeddings are processed through a stack of transformer encoder blocks. The resulting features are passed to a convolutional decoder, which upsamples and refines the prediction map. Colored arrows indicate key operations. The right panel illustrates the internal components of the Multi-Head-Self-Attention mechanism, including the scaled dot-product attention.

### *3.4.3. Swin Transformer backbone combined with UperNet decoder*

To facilitate effective spatial learning without the computational cost of global self-attention, the Swin Transformer (ST) introduces a hierarchical ViT architecture with window-based multi-head self-attention (W-MSA) computed in fixed, non-overlapping windows (Figure S3) (Liu et al., 2022). Unlike the standard ViT (Section 3.4.2), which applies attention across the entire image, W-MSA restricts attention to local windows, substantially reducing computational complexity (Xu et al., 2021; He et al., 2022). Details on the architecture design and layer configuration are provided in the Supplementary Material (Figure S3). To optimize the ST model for our feature extraction task, we adopt the same strategy used for the Vanilla ViT (Section 3.4.2) and initialize it with ImageNet-pretrained weights. In the fine-tuning, the gradients in the first two backbone stages were frozen to mitigate gradient noise, improve model performance, and reduce training data needs (Parthasarath et al., 2024). To ensure a consistent comparison, all training hyperparameters were the same as those used for the Vanilla ViT, including the optimizer configuration, learning rate schedule, loss function, and data augmentation and normalization strategy. This controlled setup ensures that any observed differences in performance can be attributed directly to architectural improvements rather than to variations in training conditions.

#### *3.4.4. SegFormer*

The development of SegFormer models is based on the integration of Transformer-based learning with convolutional inductive biases (Xie et al., 2021). Its architecture, illustrated in Figure S4, consists of two main components: the Mix Transformer (MiT) encoder and a lightweight decoder head. The MiT encoder adopts a hierarchical design similar to CNNs, where patch embedding and overlapping patch merging progressively reduce spatial resolution while enriching semantic abstraction (Yang et al., 2022). This structure allows the encoder to capture both local detail and global context efficiently. The decoder head fuses encoder outputs by projecting them into a common embedding dimension, producing a high-resolution semantic segmentation map. Additional details on the SegFormer architecture for algal bloom segmentation, including the layer configuration applied, are provided in the Supplemental Material (Figure S4). Following the Vanilla ViT (Section 3.4.2), we relied on pretrained weights loaded into the encoder to support more effective segmentation, while the decoder head was initialized from scratch. All training hyperparameters, including the learning rate, optimizer, and learning rate scheduler, were kept consistent with the configurations used in the previous models. The loss function was also based on the same composite formulation, with the addition of a distance-based weighting map (originally introduced in the ResUNet model, Section 3.4.1) incorporated into the loss function.

#### *3.4.5. Prithvi Foundational Model fine-tuning*

In recent years, Foundation Models (FMs) have emerged to address the challenge of limited labeled data in remote sensing (Xiao et al., 2024). Their development begins with a Self-Supervised Learning (SSL) framework trained on a large unlabeled dataset to learn transferable and generalizable representations, allowing the model to adapt to new tasks with minimal fine-tuning and no changes to the architecture (Choi et al., 2025). For instance, Prithvi, developed through a collaboration between IBM and NASA, was the first FM designed for Earth Observation, pre-trained on over 1 TB of Harmonized Landsat-Sentinel-2 (HLS) imagery using the Masked Autoencoder (MAE) architecture (Jakubik et al., 2023; Szwarcman et al., 2025). MAE is a ViT-based model designed to reconstruct original signals from partial observations (He et al., 2021). During training, the model divides the input image into non-overlapping patches, where a random subset of these tokens is masked, and the encoder learns to generate contextualized latent representations from the unmasked tokens by applying consecutive attention blocks. Once the encoder processes the input and masks the latent sequence, the decoder reconstructs the masked patches using the learned masked tokens and the encoder's output. The loss function compares the predicted pixel values of the masked regions with their ground-truth counterparts (Cong et al., 2023).

In our study, we used Prithvi V2 (600M) model (Szwarcman et al., 2025) as the backbone ViT encoder to generate attention maps (Figure S5). To fine-tune the FM model to our downstream task, the encoder weights were frozen (non-trainable) in the first 20 epochs to retain the generalization capabilities during training. Next, rather than using the original MAE decoder for reconstruction, we integrated the decoder from the SegFormer model (Section 3.4.4), which efficiently upsampled the multi-scale features extracted by the encoder layers. This hybrid architecture combines the self-supervised representations learned by Prithvi with the lightweight decoding capability of SegFormer, enabling pixel-level segmentation, even under limited supervision. The training was performed using the AdamW optimizer, a StepLR scheduler, and cross-entropy loss. Following the approach used in the ResUNet and SegFormer, we pre-computed the weight map for each ground truth mask, which was then combined into the loss function, to force the network to learn the small borders from algal blooms during backpropagation.

### *3.5. Accuracy evaluation*

Model evaluation was performed using the 2024 independent validation dataset (12,255 patches; Section 3.3) by comparing predictions with ground-truth features at pixel-level and patch-level metrics. For each ground-truth patch, a 16-pixel border was excluded to reduce susceptibility to prediction artifacts (Hu et al., 2023b). At the pixel level, predictions were evaluated by computing the number of true positives (TP), false positives (FP), true negatives (TN), and false negatives (FN). From this, standard classification metrics were derived, including precision ($\text{precision} = \text{TP}/(\text{TP} + \text{FP})$), recall ($\text{recall} = \text{TP}/(\text{TP} + \text{FN})$), and F1-score (also known as the Dice coefficient) ($\text{F1} - \text{score} = (2 \times \text{precision} \times \text{recall})/(\text{precision} + \text{recall})$). The F1-score is defined as the harmonic mean of precision and recall, capturing both the accuracy of positive predictions (precision) and the model's sensitivity in detecting actual bloom pixels (recall) (Terven et al., 2025). Additionally, omission and commission errors were calculated to assess model reliability (Foody, 2002). Omission error quantifies the proportion of actual bloom pixels that were incorrectly predicted as non-bloom, highlighting under-segmentation or missed detections, and is calculated as $\text{omission} = (\text{FN}/(\text{TP} + \text{FN})) \times 100$. While the commission error, calculated as $c\text{ommission} = (\text{FP}/(\text{TP} + \text{FP})) \times 100$, measures the proportion of predicted bloom pixels that are non-bloom, indicating over-segmentation or false detections (Martins et al., 2022; Salleh, 2023). Patch-level validation metrics were also performed in terms of algal bloom area coverage within patches. This approach evaluates the influences of the proportion of bloom pixels relative to total patch size and allows assessment of the model’s ability to preserve bloom extent and spatial structure across regions. Because the ground-truth was not obtained from field measurements but generated from a semi-automatic approach, such metrics should be regarded as self-consistent (Hu et al., 2023b). Finally, the impact of radiometric uncertainty on model performance was evaluated by introducing distinct levels of $R_{rs}$ noise derived from AQUAVis validation, with full methodological details described in Appendix S3.

## 4. Results

### *4.1. Benchmarking deep learning architectures for algal bloom segmentation*

Table 1 illustrates normalized confusion matrices for the algal bloom segmentation. Across all models, both true negative (background correctly classified) and true positive (algal bloom correctly classified) rates are generally high, reflecting robust performance in distinguishing bloom presence. Among them, SegFormer achieved the highest true positive rate (92.07%), demonstrating strong capability in detecting bloom regions. However, this came at the cost of a lower true negative rate (85.83%), indicating a higher tendency toward false bloom detections.

The MAE (Prithvi) model exhibited the worst bloom classification performance, with a true positive rate of 66.63%, suggesting limited sensitivity and a tendency to miss portions of bloom events. ResUNet and Swin Transformer both showed well-balanced performance. ResUNet achieved a true negative rate of 90.27% and a true positive rate of 90.21%. Swin Transformer reached a true negative rate of 90.23% and a true positive rate of 87%, demonstrating reliable performance for both classes.

Table 1. Per-class confusion matrices (in percentage) for algal bloom segmentation using five deep learning models: ResUNet, Vanilla ViT, Swin Transformer, SegFormer, and MAE (Prithvi). Each row shows the proportion of correctly and incorrectly classified pixels for bloom and background classes, normalized independently per class to account for dataset imbalance. Best performance values for each metric are highlighted in bold.

| Model | True-Negative | False-Positive | False-Negative | True-Positive |
|---|---|---|---|---|
| **ResUnet** | 90.27 | 9.73 | 9.79 | 90.21 |
| **Vanilla ViT** | 87.56 | 12.44 | 15.39 | 84.64 |
| **Swin Transformer** | 90.23 | 9.77 | 13.00 | 87.00 |
| **SegFormer** | 85.83 | 14.17 | **7.93** | **92.07** |
| **MAE (Prithvi)** | **92.06** | **7.94** | 33.37 | 66.63 |

Additional error metrics are shown in Table 2, where among the ViT-based models, Swin Transformer delivered the best F1-score (0.57), with also improved performance in terms of commission errors (57.51%). SegFormer achieved the highest recall (0.92) but also the highest commission error (64.96%) and lowest precision (0.35), indicating a strong tendency to overpredict bloom pixels. MAE (Prithvi) showed the lowest recall (0.67) and highest omission error (33.26%), reflecting poor sensitivity to bloom regions among the evaluated models. Vanilla ViT presented intermediate performances, with an F1-score of 0.50 and a trade-off between omission and commission. For comparison, ResUNet achieved the highest precision (0.43) and F1-score (0.58). Despite the relatively high recall and precision values in Table 1, all F1-scores (Dice coefficients) remained below 0.60, likely reflecting the fine-scale, fragmented spatial structure of algal blooms. In such conditions, even minor localization errors can greatly reduce overlap-based metrics like the Dice coefficient. Moreover, these results stem from out-of-distribution inferences, where performance drops relative to in-distribution tests are common and well-documented. To further assess whether the observed errors were due to true misclassifications, we included an additional analysis in Table S1 by applying a one-pixel buffer around each ground truth mask to estimate a confidence interval for the reported metrics. This simple adjustment led to a marked improvement in performance across all models. From Table S1, the Swin Transformer showed the closest performance to ResUNet, with an F1-score of 0.70, and commission and omission errors of 25.28% and 28.39%, respectively. ResUNet, for instance, showed an improvement in the balance between omission (29.57%) and commission (31.46%) errors, leading to a F1-score of 0.70.

Table 2. Evaluation metrics for algal bloom segmentation using five deep learning models: ResUNet, Vanilla ViT, Swin Transformer, SegFormer, and MAE (Prithvi). Metrics include recall, precision, F1-score (Dice coefficient), commission error (false positives), and omission error (false negatives). Best performance values for each metric are highlighted in bold.

| Metrics | ResUNet | Vanilla ViT | Swin Transformer | SegFormer | MAE |
|---|---|---|---|---|---|
| **Recall** | 0.90 | 0.84 | 0.87 | **0.92** | 0.67 |
| **Precision** | **0.43** | 0.36 | 0.42 | 0.35 | 0.41 |
| **F1-Score (Dice)** | **0.58** | 0.50 | 0.57 | 0.51 | 0.51 |
| **Commission (%)** | **56.59** | 64.05 | 57.51 | 64.96 | 59.00 |
| **Omission (%)** | 9.88 | 15.52 | 13.12 | **7.99** | 33.26 |

### *4.2. The impact of bloom-affected area on model performance errors*

Because algal blooms often appear as small, fragmented features, their coverage within a patch strongly affects detection and delineation. Figure 6 analyzes omission and commission errors by bloom proportion, while Figure 7 shows segmentation results across different bloom patterns. Following the approach described in Section 4.1, we recalculated the metrics presented in Figure 6 by incorporating a one-pixel confidence buffer around the ground truth labels. The resulting performance metrics are summarized in Table S2. From Figure 6, omission error tends to dominate in patches with low bloom coverage, while commission error becomes more prominent as bloom coverage increases. For example, in the lowest algal proportion (0–0.1%), all models exhibit high omission rates, ranging from 44.97% (Vanilla ViT) to 64.36% (ResUNet), indicating some difficulty in detecting small, sparse, and fragmented bloom features. From Table S2, omission errors were notably reduced, ranging from 19.20% to 28.53%. Conversely, commission error remains low in this range, typically below 5%, reflecting low false positive rates when bloom is nearly absent.

As the bloom proportion increases, the models become more sensitive, reducing omission errors but at the cost of increasing commission errors. An exception to this trend is the MAE model, where omission errors remained high even under high bloom conditions. SegFormer showed aggressive bloom detection in mid-to-high bloom bins (> 45%), with omission errors as low as 16.34% but commission errors rising to 73.56% (48.11%, Table S2), suggesting a tendency toward over-segmentation. In contrast, both ResUNet and Swin Transformer exhibit relatively stable performance, maintaining a balance between omission and commission errors across a wide range of bloom proportions. The highest commission error for ResUNet was 62.43%, observed at bloom proportions above 85%, whereas the Swin Transformer reached a peak commission error of 70.67% in the same bloom proportion range. After applying a one-pixel confidence buffer around the ground truth labels, the maximum commission error was substantially reduced to 50.47% for Swin Transformer. At moderate bloom proportions (25–45%), ResUNet exhibited omission and commission errors of 27.35% and 32.35%, respectively, while the Swin Transformer showed similar performance with omission and commission errors of 25.12% and 33.16% (Table S2). This improvement indicates that a considerable portion of the initial errors arose from minor spatial misalignments, rather than actual misclassification of bloom areas.

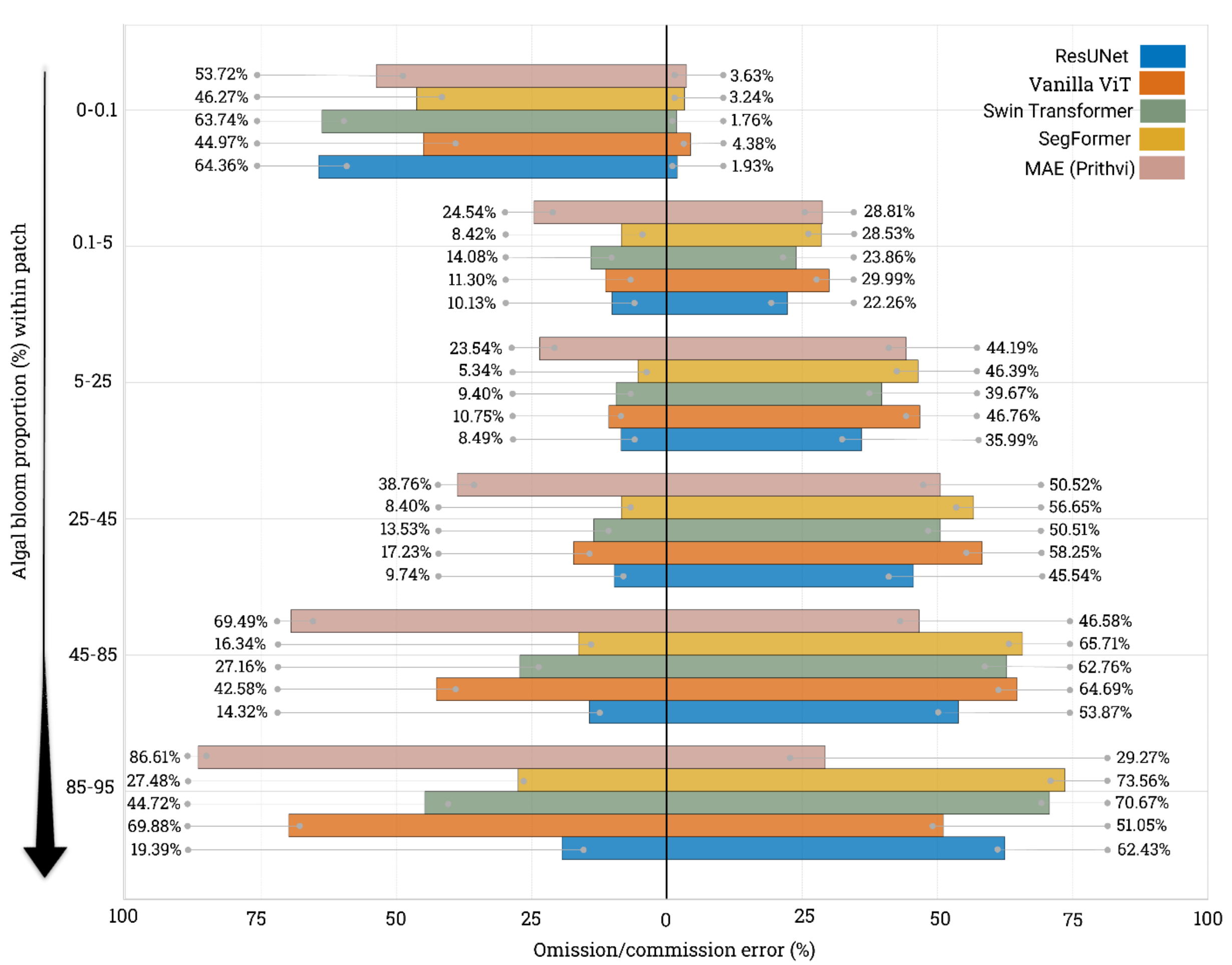

Figure 6. Sensitivity analysis of each DL model's accuracy to the algal bloom area proportion. Results shown for six proportion groups composed of 12,255 256x256 30-meter patches per group.

Figure 7 shows segmentation examples for different algal bloom proportions within the patch, ranging from sparse to extensive bloom coverage. Each column corresponds to a different test patch, with the two top rows showing the true-color and false-color composites and the subsequent rows displaying model predictions overlaid with ground truth. Green indicates correctly classified pixels (true positives), yellow represents commission error (background classified as bloom), and red denotes omission error (bloom missed by the model). Overall, all models successfully identified algal bloom occurrence and demonstrated an ability to learn both dense and subtle bloom patterns. In patches with low bloom coverage (first column), Swin Transformer, SegFormer, and ResUNet achieved higher proportions of correctly classified pixels, as also reflected in the bottom-row bar plots. In contrast, MAE and Vanilla ViT struggled to detect subtle bloom structures, resulting in higher omission errors. As bloom proportion increases, all models show improved detection, but with a concurrent rise in commission error (yellow), particularly in Vanilla ViT and SegFormer, which tended to over-segment bloom regions. This pattern is most evident in the rightmost column, where SegFormer captures dense bloom areas but substantially overpredicts into surrounding non-bloom waters. Swin Transformer, on the other hand, maintained relatively low commission errors even at the highest algal bloom proportions. A closer examination of the spatial distribution of the predicted masks relative to the ground truth reveals that the majority of high commission errors observed in Figure 6 and Table 2 stem primarily from the buffered nature of the predictions around the bloom boundaries, rather than from random or spatially inconsistent false positives. Additional examples in Figure S6 further illustrate these patterns across diverse spatial configurations of bloom events.

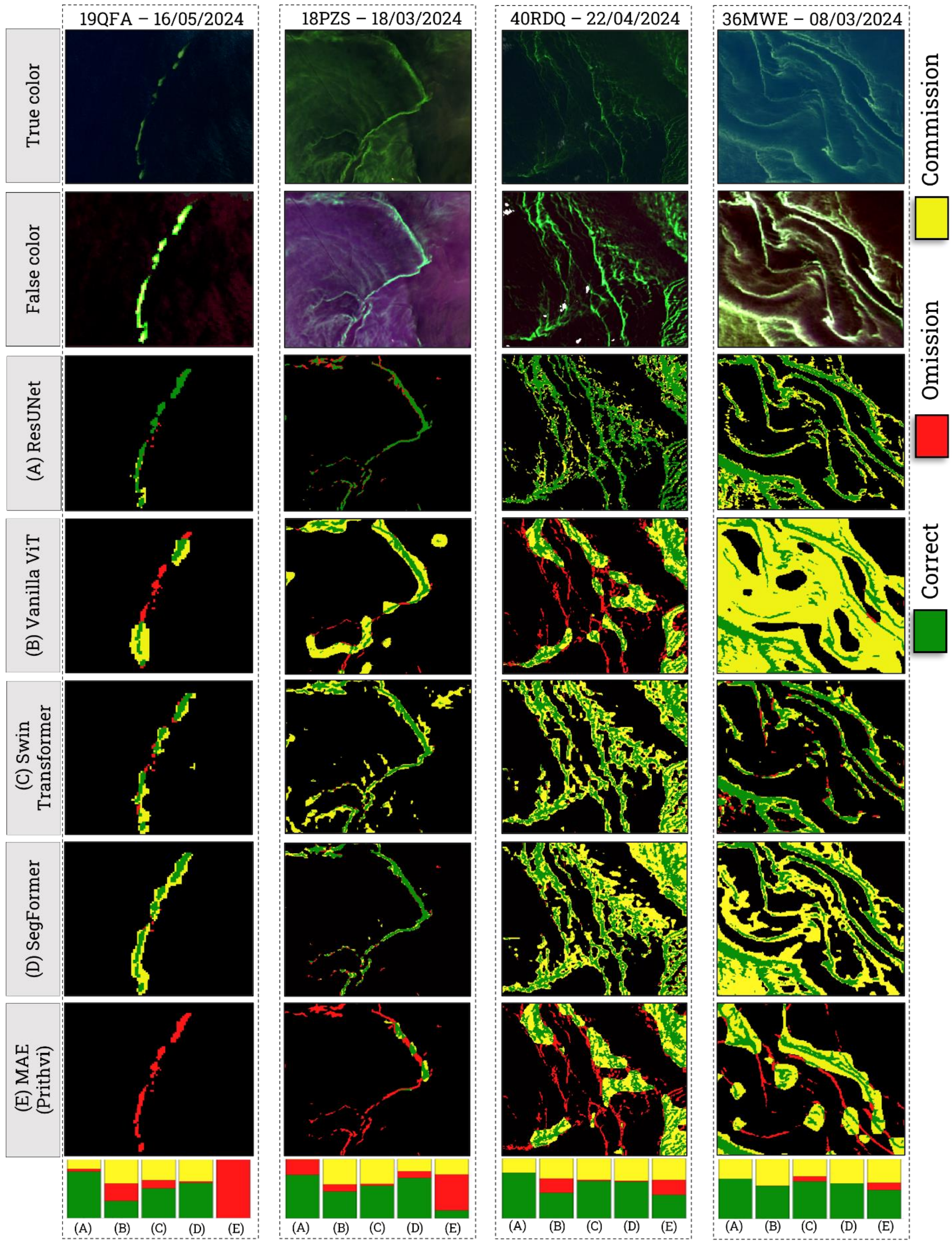

Figure 7. Examples of algal bloom results for proposed models. The top two rows show the true and false color (red, NIR, green) compositions of the algal bloom occurrences, while the remaining

rows illustrates the segmentation results and algal bloom reference masks together, showing correctly classified pixels (green), pixels that should have been classified as algal bloom but were not (red), and pixels that were incorrectly classified as blooms (yellow). The bar plots in the bottom row show the correct, omission, and commission proportions for each model. The location of each patch example is indicated by its MGRS tile ID, with corresponding tile centroids as follows: 19QFA (67.53°W, 18.49°N), 18PZS (71.76°W, 10.34°N), 40RDQ (56.54°E, 26.62°N), 36MWE (33.49°E, 0.50°S).

### *4.3. Evaluating model performance under diverse environmental conditions*

Based on the previous analysis, Swin Transformer demonstrated the best overall performance among the ViT-based evaluated models, and in this section, we examine its behavior under challenging real-world conditions. First, to demonstrate the model mapping capability under verified events, we utilized a bloom occurrence dataset compiled by Gernez et al. (2023) and Qi et al. (2020, 2025). As detailed in Table S3, 80 occurrences were cataloged between 2015 and 2022 across coastal waters in 21 countries, spanning 27 species, 20 genera, and 5 classes, from which Swin Transformer accurately identified 52. Figure 8 illustrates representative detections across diverse coastal environments, years, morphologies, and taxa. These examples encompass eight different classes, including both micro- and macroalgae, and highlight how the model successfully captured complex spatial structures such as filamentous streaks, patchy aggregations, and large-scale surface accumulations. Performance remained robust across optically complex environments. For example, dense *Microcystis* blooms in Lagoa dos Patos are accurately mapped despite elevated turbidity and reduced water–bloom spectral contrast, conditions that typically confound threshold-based or index-driven approaches. Similarly, filamentous patterns observed in coastal and open-ocean settings (e.g., Baltic Sea and Arabian Sea cases) are well reproduced, indicating sensitivity to mesoscale and sub-mesoscale features.

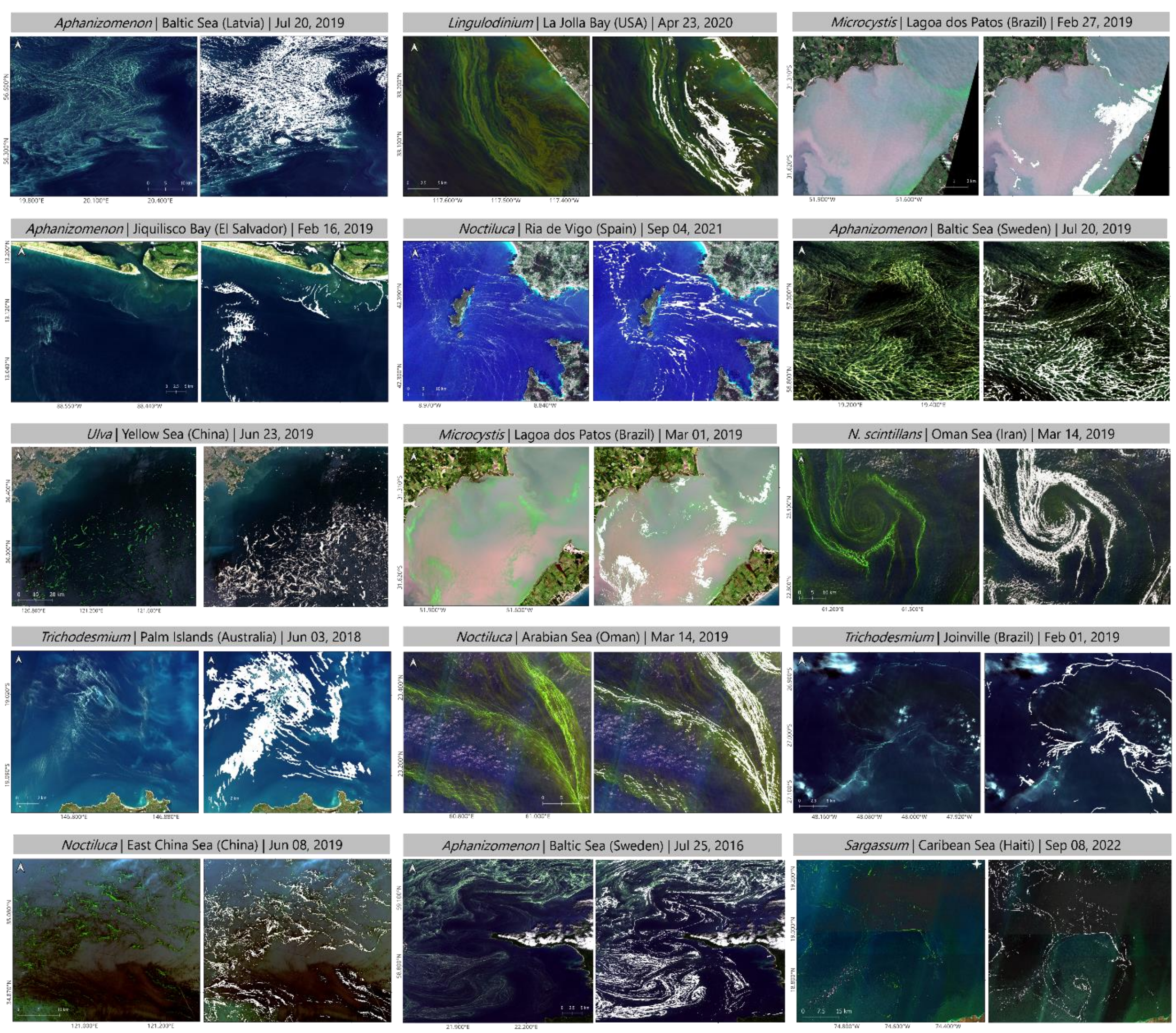

Figure 8. Qualitative evaluation of Swin Transformer segmentation across a wide range of real-world algal bloom events. Each panel presents a paired visualization consisting of an RGB composite (left) and the corresponding model-derived bloom mask (right), where detected algal presence is shown in white. The examples span multiple geographic regions, acquisition dates, water types, and bloom-forming taxa, highlighting the model's ability to generalize across diverse environmental and optical conditions.

Figure 9 provides a detailed visual assessment of Swin Transformer's segmentation performance across the distinct OWTs described in Section 2.2 (Figure S1). Classification maps from the other models are also provided for comparison in Figures S7 and S8 for each evaluated scenario. For OWT 1 (high chlorophyll a), ST successfully captured the complex bloom structures, as evidenced by the strong spatial alignment with the ground truth and minimal omission or commission errors in the selected zoom regions. This indicates good sensitivity to dense bloom features typically associated with eutrophic waters. In OWT 2 (high TSS), while some linear bloom patterns were correctly detected, the highlighted areas showed some commission errors (yellow). However, it is important to highlight that the model did not misclassify high TSS regions as algal bloom, indicating good discrimination between sediment and bloom signals. For clear waters (OWT 3), the model maintained strong performance, with accurate detection of faint bloom filaments and minimal misclassification. The subtle bloom structures presented in the oligotrophic environment were well captured, demonstrating the model's capacity to generalize across low signal environments. As shown in Figure S7, all ViT-based models were visually capable of identifying algal blooms across the different aquatic environments. Vanilla ViT and MAE, as expected, exhibited smoother predictions and failed to capture finer algal filaments, particularly in scenes with complex spatial patterns. In contrast, SegFormer demonstrated strong performance, successfully detecting bloom features even in challenging turbid conditions.

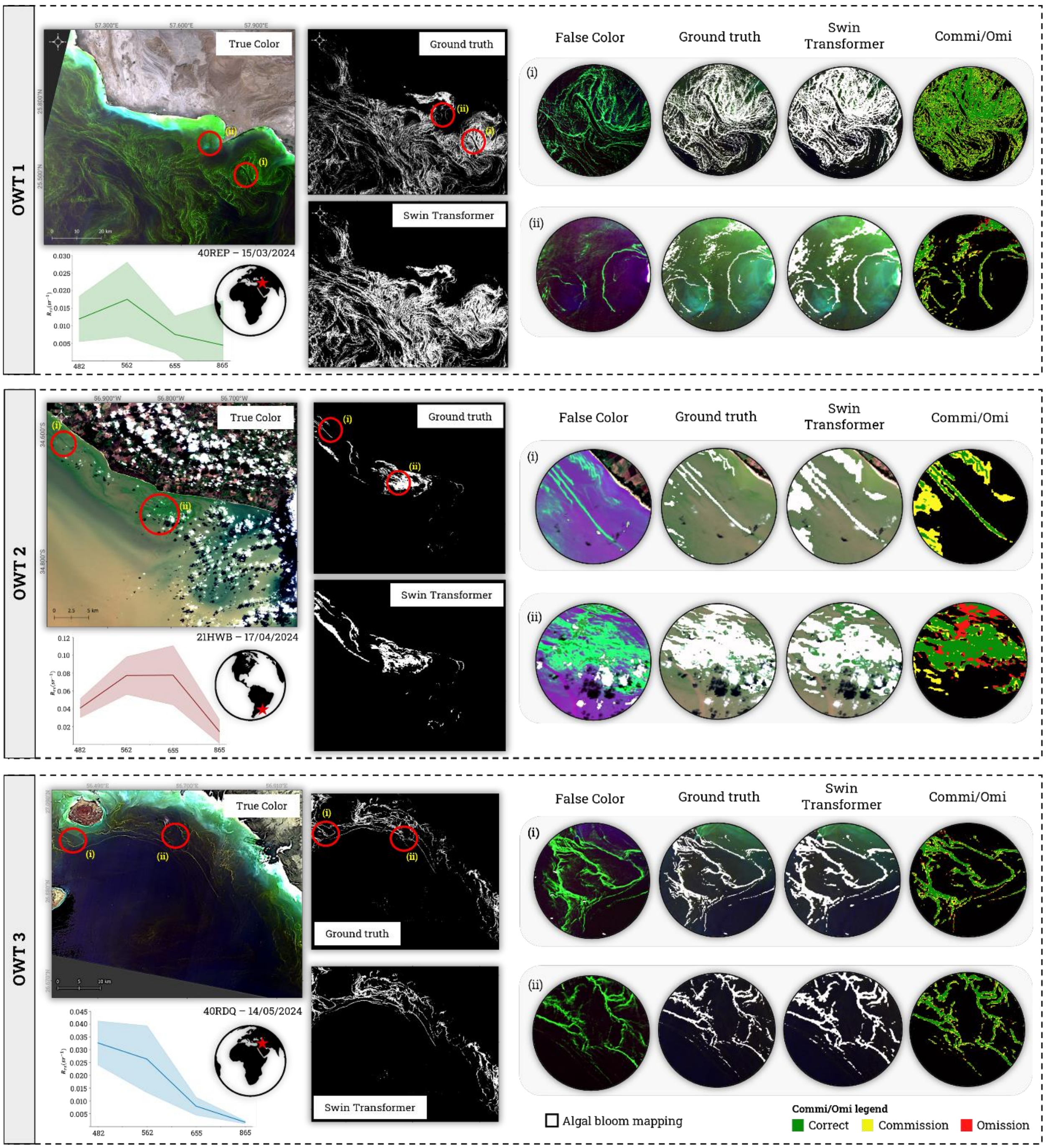

Figure 9. Segmentation examples for three OWTs using Swin Transformer. For each OWT, the true color composite (left) shows the full scene, followed by the ground truth mask and the ST prediction, with red circles marking selected regions of interest. These regions are highlighted in the circle images and include: false color image (red, NIR, green), ground truth, ST output, and error overlay. Reflectance curves are also provided for each OWT. The location of each example is indicated by its MGRS tile ID, with corresponding tile centroids as follows: 40REP (57.55°E, 25.72°N), 21HWB (56.40°W, 34.83°S), 40RDQ (56.54°E, 26.62°N).

Figure 10 presents a visual evaluation of the Swin Transformer model's performance under two atmospheric stress scenarios commonly encountered in coastal monitoring: (A) high-altitude cloud cover and (B) intense glint interference. For each case, Figure 10 shows the true-color composite, the corresponding ground truth mask, and the ST prediction. Cloud and shadow masks are plotted to highlight the cloud/shadow coverage masked by Fmask and the remaining interference that directly affected the model's performance. In Figure 10A, despite substantial occlusion by thin clouds that were not masked by FMask, Swin Transformer effectively captured the main bloom structures with notable spatial coherence. The omission and commission error maps in the zoomed-in regions reveal that, although the model exhibited buffered overprediction of bloom boundaries (as indicated by the yellow areas), the majority of bloom filaments were correctly segmented, demonstrating low sensitivity to thin cloud interference. In the second case (Figure 10B), strong glint introduced spectral distortions that typically pose a significant challenge for optical remote sensing. However, the DL model effectively suppressed much of the glint-induced noise and successfully retrieved the bloom patches. The zoomed-in regions highlight this performance, showing minimal omission errors and no notable false positives attributable to glint. For both evaluated scenarios, the classification maps produced by ResUNet, ViT, MAE, and SegFormer are shown in Figure S8. Among them, ResUNet exhibited the most precise delineation, closely matching the ground truth, followed by SegFormer. Consistent with previous analyses, Vanilla ViT and MAE showed smoother and more diffuse outputs. Despite these variations, none of the models appeared highly sensitive to thin clouds or glint artifacts, successfully delineating algal bloom features in their presence.

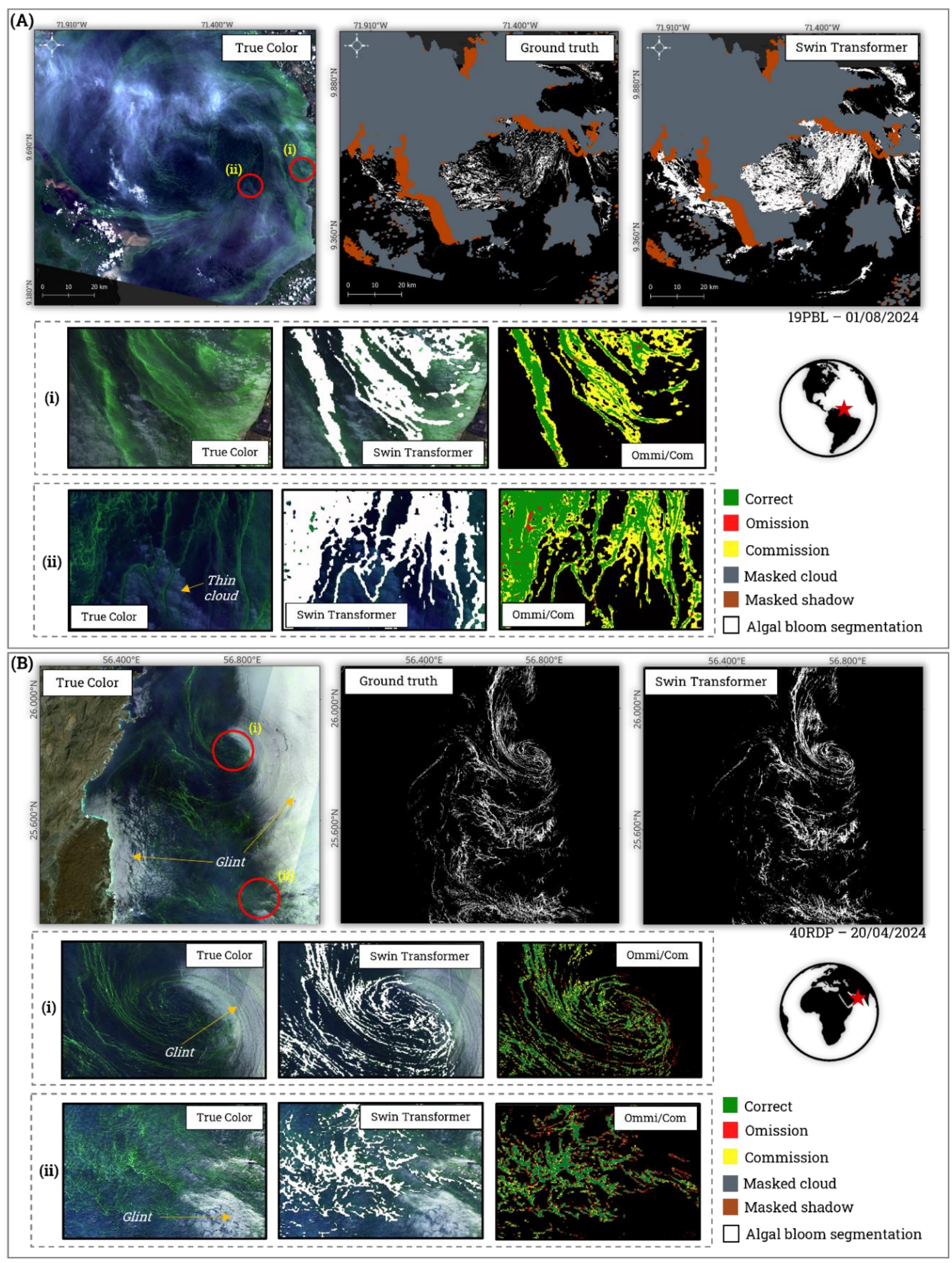

Figure 10. Performance of the Swin Transformer model under challenging atmospheric conditions: (A) is a case dominated by high-altitude clouds, and (B) is a case affected by strong glint + cloud.

For each scene, true-color composites, ground truth masks, and Swin Transformer predictions are presented. The cloud and shadow masks are also plotted. Insets below each case provide zoomed-in visual comparisons between true-color imagery, ST predictions, and omission/commission outputs for selected regions of interest. Segmentation results for ResUNet, ViT, MAE, and SegFormer models are provided in Figure S8. The location of each example is indicated by its MGRS tile ID, with corresponding tile centroids as follows: 19PBL (71.23°W, 9.44°N), 40RDP (56.55°E, 25.72°N).

### *4.4. Beyond thresholds: evaluating traditional vs. deep learning approaches for algal bloom mapping*

To assess algal bloom mask quality under challenging imaging conditions, we compared the Swin Transformer model with traditional spectral index-based methods, focusing strictly on spatial delineation accuracy without discounting the fundamental value of indices as physically interpretable proxies. Figure 11 shows a four-day temporal evaluation of NDVI, FAI, and Swin Transformer performance. For the index-based methods, a fixed threshold of zero was applied uniformly across all dates to reflect standard operational practice. While scene-by-scene threshold optimization might yield higher performance for individual images, such an approach is rarely feasible at large scales. Consequently, this comparison represents a realistic operational baseline rather than the theoretical maximum performance of each index; the qualitative results should be interpreted within this context. Visual inspection on April 7 and 20 (Figure 11, A and B) indicates algal bloom occurrences in the true-color composites, where all methods successfully detected their presence, with NDVI and FAI producing some noisy responses in glint-affected areas. The contrast becomes more evident on April 22 and 27 (Figure 11, C and D), when no blooms are present but strong sun-glint dominates the scene. Here, NDVI and FAI falsely classified large glint-affected regions as blooms, highlighting their sensitivity to spectral distortions. In contrast, the Swin Transformer effectively distinguished glint artifacts from true bloom signals, avoiding false positives and demonstrating greater robustness under challenging optical conditions without the necessity of conservative masks. We also evaluated the sensitivity of FAI to different thresholds, and the examples in Figure S9 further demonstrate that FAI is highly sensitive to small threshold variations, with different dates exhibiting distinct “optimal” thresholds, underscoring the difficulty of developing an operational and scalable workflow for bloom segmentation using fixed index thresholds under varying glint, adjacency, and atmospheric conditions.

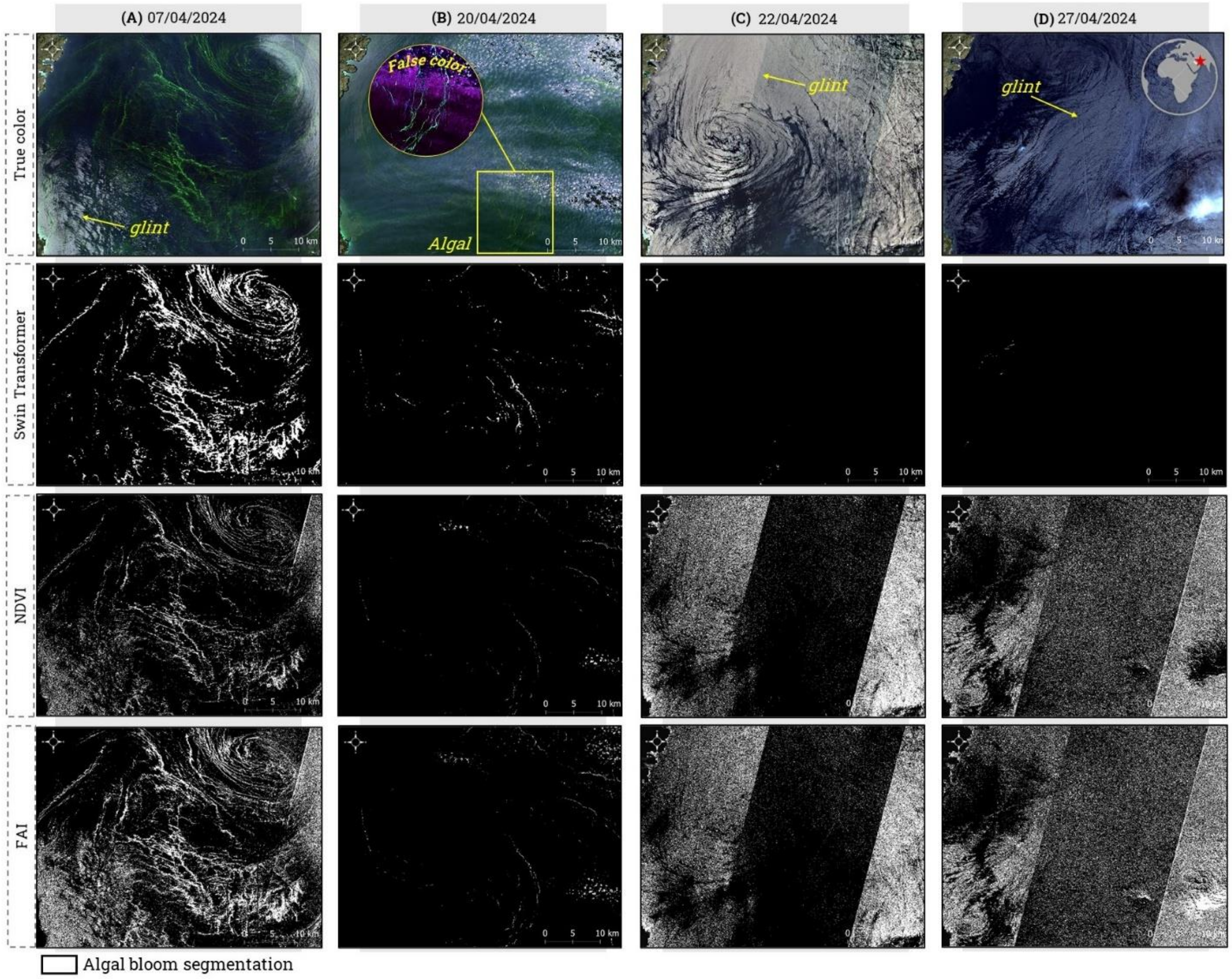

Figure 11. Temporal comparison of algal bloom detection using Swin Transformer, NDVI, and FAI across four consecutive cloud-free observations within the same month. Each column represents a different date, while rows correspond to: (1) true-color composites (top), (2) Swin Transformer predictions, (3) NDVI-based masks, and (4) FAI-based masks. Glint-affected regions are indicated with yellow arrows in the true-color images. Spectral index masks were generated using a threshold of 0. MGRS tile ID of the image scene is 40RDP (56.55°E, 25.72°N).

In the second scenario, Figure 12 highlights the impact of spatial resolution on algal bloom detection by comparing the medium-resolution Swin Transformer predictions (30 meters) against coarser MODIS-derived daily bloom masks at 1 km resolution (Dai et al., 2023). While MODIS provides useful large-scale information, its coarse spatial detail often fails to capture the fine-scale complexity of blooms, especially in nearshore regions that are commonly masked to avoid land pixels. The zoom-in areas in Figure 12 illustrate how the ViT-based model delineates small bloom filaments and edges closely aligned with visible patterns in the true color imagery. In contrast, MODIS mask appears (red mask in Figure 12) fragmented and spatially generalized, missing smaller bloom patches and overestimating presence in uniform areas. Quantitatively, Swin Transformer predicted a bloom extent of 1,225.57 km², compared to 1,298 km² from MODIS. However, MODIS estimates should be interpreted only as a lower bound, since the "missing" positives from coarse resolution could represent more than 81% of the bloom area detected at 30 m with AQUAVis. Despite the overall area extent, the finer spatial detail of AQUAVis provides more accurate delineation of bloom patterns, enabling the detection of subtle structures that are vital for early-warning systems, ecological assessments, and informed management decisions.

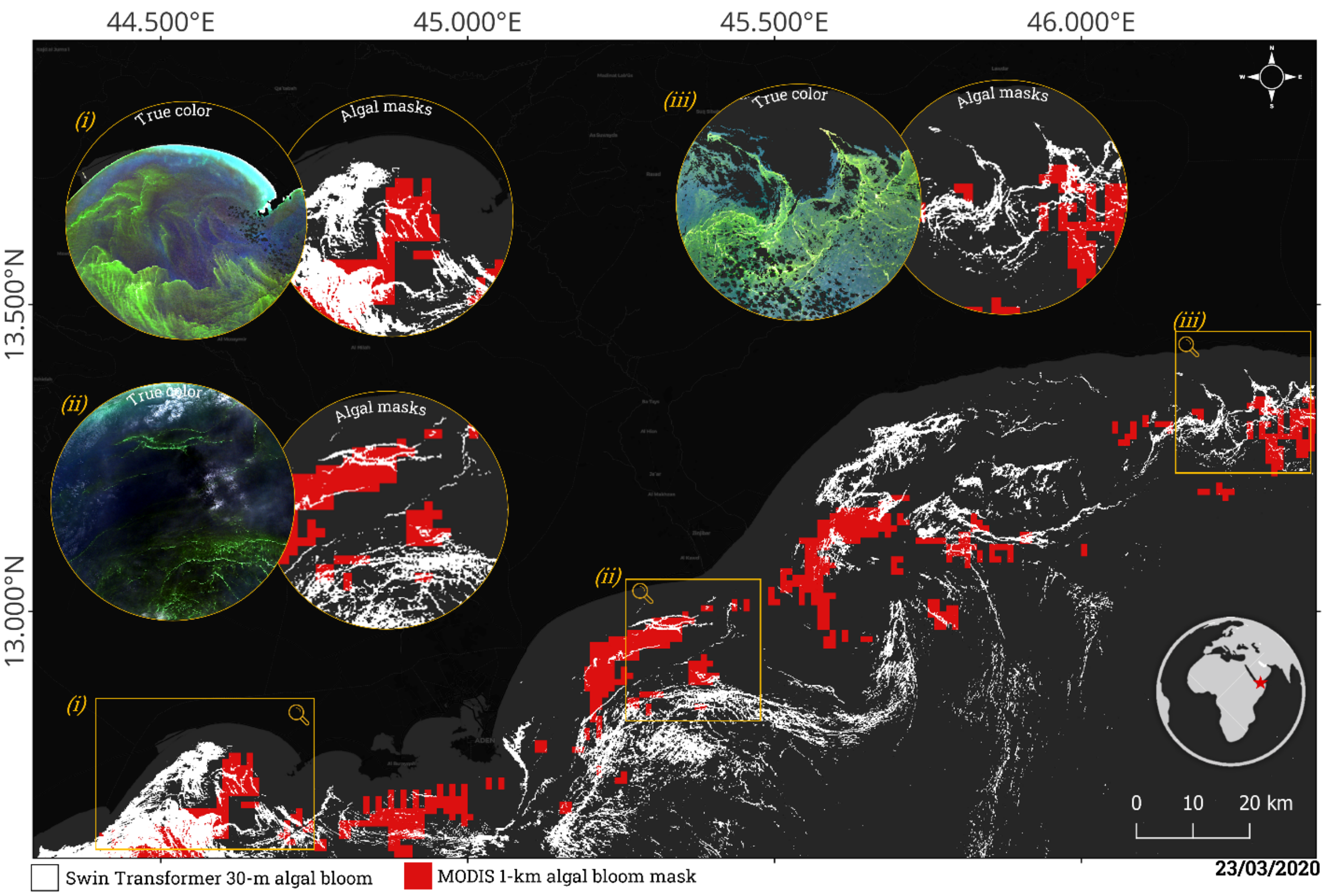

Figure 12. Comparison of algal bloom detection using medium-resolution Swin Transformer predictions (30 m, in white) and MODIS-derived daily bloom masks (1 km, in red) for a nearshore region (<50 km from the coast). The main panel displays the spatial distribution of both masks, while the insets (i–iii) highlight specific areas with corresponding true color composites for visual reference.

## *5. Discussion*

### *5.1. From pixels to patterns: Rethinking algal bloom mapping with deep learning and transformers*

Optical remote sensing provides broad spatial and temporal coverage for monitoring algal blooms, but variability in water constituents and atmospheric conditions challenges global model generalization, with most existing approaches remaining region-specific (Jeon, 2023; Hu et al., 2023b, 2025). Here, we demonstrated a globally applicable algal bloom mapping by integrating multi-sensor satellite data with five deep learning architectures, including Vision Transformers. All evaluated models showed a strong ability to detect algal blooms in nearshore waters, with different levels of precision regarding minor details. Among ViT-based models, Swin Transformer showed the best delineation of local structures (F1-score of 0.570), accurately capturing bloom extent in complex aquatic environments (Figure 7). Validation against 2024 data and 2015–2022 occurrences demonstrates that the model's spectral-spatial representations generalize across geographic regions, acquisition dates, water types, and bloom-forming taxa, highlighting the model's ability to generalize across diverse environmental and optical conditions, and how the performance relies on bloom morphology and optical contrast rather than year-specific anomalies.

Comparisons between CNN- and ViT-based models highlighted the role of architectural design. ResUNet achieved the highest performance (F1 = 0.585) and maintained balanced omission and commission errors (29.57% and 31.46%, respectively), effectively detecting blooms from fragmented patches to high-coverage events (Figure 6). While CNN inductive biases help identify fine textures (Wang et al., 2019, 2021; Martins et al., 2022), they may struggle in unseen scenarios due to their limited receptive field (Hu et al., 2023a). As detailed in section 3.4.2, ViTs improve generalization through global self-attention, which captures long-range dependencies by computing pairwise relationships across all spatial tokens, moving beyond the local receptive fields of traditional CNNs (Xu et al., 2021; Dosovitskiy et al., 2021). However, this global focus can lead to the over-smoothing of pixel-level boundaries, a common artifact in Vanilla ViT and MAE architectures (Figures 7, S7, S8). To improve pixel-level segmentation, Swin Transformer and SegFormer combine the global context modeling of self-attention with the spatial locality and efficiency of convolution (Xie et al., 2021; Liu et al., 2021). Both models showed reduced omission and commission errors (Figure 6) and better captured fine-grained spatial details. Swin Transformer, for example, captures local context by applying self-attention within windows, focusing on spatially localized features (Figure S3). Furthermore, due to architectural design, the models also differed in computational complexity. During training, ResUNet, with 29.1M parameters, achieved the fastest per-epoch time (1.7 min), followed by SegFormer (2.5 min with 44.6M parameters) and Swin-T (3.3 min with 32.6M parameters). In terms of inference, Vanilla

ViT (24M) and SegFormer were the most efficient, requiring 16s and 58s for a full scene, while ResUNet required 15 min for full-scene inference, as convolutional encoder–decoder architectures scale less efficiently with large spatial inputs. Swin Transformer offered the best balance between accuracy and computational cost with an inference runtime of 4 min 28 s. The MAE (Prithvi) model, with 305M parameters, was by far the heaviest and exhibited the largest computational footprint during training and inference.

### *5.2. Toward Scalable algal bloom mapping across diverse aquatic environments*

The model's performance was compared against algal bloom occurrences and evaluated under diverse atmospheric and environmental conditions. Across Optical Water Types (OWTs), Swin Transformer and other ViT-based models reliably delineated bloom structures, even in turbid or optically clear waters (Figure 9). Similar behavior was observed under high thin cloud and glint conditions, where Swin Transformer consistently delineated bloom structures (Figure 10 and S8). While FMask filters out heavy cloud cover, residual thin clouds and glint can distort pixel spectra, also providing an ideal context to demonstrate the advantages of DL models over traditional spectral index methods (Figure 11). For instance, Hu et al. (2023b) applied a U-Net model to detect *Sargassum* in the Atlantic Ocean using MODIS imagery and compared its performance with the AFAI spectral index. Consistent with the results shown in Figure 11, they found that nearshore and cloud-adjacent pixels were frequently misclassified as *Sargassum* by the AFAI index, while the U-Net model successfully avoided these false positives, accurately identifying true algal features. As discussed in the Appendix S2 and highlighted in Figure 11 and S9, despite the DL model being trained using spectral indices as input channels, while traditional indices-only rely on absolute pixel-wise thresholds, deep learning models prioritize the relational structure between spatial elements (Blondeau-Patissier et al., 2014; Ronneberger et al., 2015). This illustrates a key advantage of DL approaches: by exploiting spectral–spatial context, they distinguish true bloom features from noise caused by atmospheric or surface effects, thereby reducing segmentation biases (Wang and Hu, 2016; Hu et al., 2019) and enhancing applicability across aquatic environments.

Medium-resolution observations were also shown to provide substantial benefits (Wang and Hu, 2021a). The results presented in Section 4.4 (Figure 12) demonstrated that our 30-meter-resolution algal bloom map revealed a much finer spatial distribution of blooms completely obscured in the 1-km MODIS-derived mask. This finding aligns with previous work by Wang and Hu (2021b), who combined multiple high-resolution sensors to train a CNN for *Sargassum* detection. Their analysis showed that higher-resolution imagery consistently captured more *Sargassum* features, due to its ability to resolve small-scale structures missed by coarser sensors like MODIS. Although the signal-to-noise ratios of MSI and OLI are lower than those of dedicated ocean color sensors, both have proven effective for aquatic applications (Pahlevan et al., 2017).

### *5.3. Limitations and future work*

In our study, algal bloom distribution was characterized using various forms of representation, and a major challenge is quantifying uncertainties arising from limited in situ measurements for validating satellite-derived estimates (Hu et al., 2023; Qi et al., 2023). As noted by Wang and Hu (2021b), even with high-resolution imagery, uncertainties persist, reflecting the difficulty of establishing an objective ground truth. We used a semi-automatic labeling approach combining visual inspection, spectral indices, and validated unsupervised detection methods (Wang and Hu, 2015), a widely accepted alternative for DL performance assessment (Martins et al., 2022; Hu et al., 2023b; Qi et al., 2025). Initial index-derived masks were extensively refined to create a final labeled masks that reflect visually identifiable bloom structures, independent of the absolute magnitude of any red/NIR/SWIR-based ratio. Despite uncertainties, our DL models learned general spectral–spatial patterns (Rolnick et al., 2018) and effectively delineated blooms across diverse environmental conditions, confirming that the curated dataset provided a robust foundation for operational applications (Wang and Hu, 2021; Martins et al., 2022; Hu et al., 2023; Yao et al., 2024). Multi-sensor comparisons using co-located imagery can further evaluate uncertainties (Wang and Hu, 2021b). In our study, we compared our classification with the algal bloom maps generated from MODIS (Dai et al., 2023), observing overall agreement, as illustrated in Figure 12. Another limitation arises from how algal blooms are defined since a variety of floating algae have been detected through satellite remote sensing (Qi et al.,2023), while other floating materials may also be mistaken for blooms (Mantas et al., 2011). This remains a recognized limitation, particularly with medium-resolution sensors that lack hyperspectral coverage. However, unlike semi-analytical approaches, DL models are data-driven, with the performance depending on the selection of representative training data. Future work may examine how these models handle additional floating materials.

ViTs, when trained on large datasets, can outperform state-of-the-art CNNs in various tasks (Zheng et al., 2021). However, their data-hungry nature remains a limitation, addressed through large-scale pre-training and task-specific fine-tuning (Thisanke et al., 2023). Here, this strategy was employed across all ViT-based models, where pre-training proved to be essential for making the training process computationally feasible in terms of both time and resource demands. However, the still-limited performance of the Prithvi pre-trained model (Figure 6, 7, S7, and S8) highlighted a major challenge in applying this approach to such domain-specific tasks: the limited availability of pre-trained models on domain-specific datasets. Here, the ViTs models were pre-trained on datasets different from aquatic environments in terms of spectral characteristics, spatial patterns, and optical complexity, limiting their effectiveness in such specialized applications. A key direction for future work is to include training datasets that encompass years characterized by anomalous bloom dynamics, which may produce bloom extents, intensities, and taxonomic compositions not well represented in the current dataset. Alongside this, the development of domain-aware foundational models trained on large-scale aquatic or ocean-specific datasets represents a compelling longer-term direction. Such models could integrate prior knowledge of water optics, seasonal dynamics, and sensor characteristics to improve performance and robustness in algal bloom detection across diverse aquatic systems.

## 6. Conclusions

The presented study offers a comprehensive benchmark for algal bloom segmentation in coastal waters, combining a globally labeled dataset, multi-sensor medium-resolution imagery, and state-of-the-art deep learning architectures. By evaluating convolutional and transformer-based models, we outline practical labeling strategies and demonstrate the applicability of Vision Transformers for robust bloom mapping. Despite challenges inherent to aquatic remote sensing, the models achieved strong detection accuracy and spatial delineation, with Swin Transformer excelling in fine-scale bloom identification and generalizable representation. These findings highlight deep learning sensitivity to intra-class variability and fragmentation, as well as the limitations of coarse sensors like MODIS. Beyond benchmarking, our framework advances scalable 30-m algal bloom mapping using Landsat–Sentinel data, providing tools for near-real-time monitoring. To our knowledge, no systematic coastal bloom mapping pipeline at this resolution exists. Our contributions include: (i) a semi-automated workflow for curating a global dataset; (ii) insights into strengths and limitations of advanced DL models for aquatic applications; (iii) evidence of their utility for future operational systems; and (iv) groundwork for domain-aware foundational models that improve generalization across diverse environments. More broadly, this work illustrates how AI can overcome limitations of traditional feature extraction and, combined with expanding multi-sensor observations, offers a powerful pathway for advancing ocean science and ecosystem monitoring.

## ACKNOWLEDGMENTS

We thank the support of grant #80NSSC24K1040 awarded by the NASA Early Career Investigator Program in Earth Science. We also thank NASA and ESA for providing access to Landsat-8/9 and Sentinel-2 A/B/C satellite data.

# Supplementary Material

## Tables and Captions

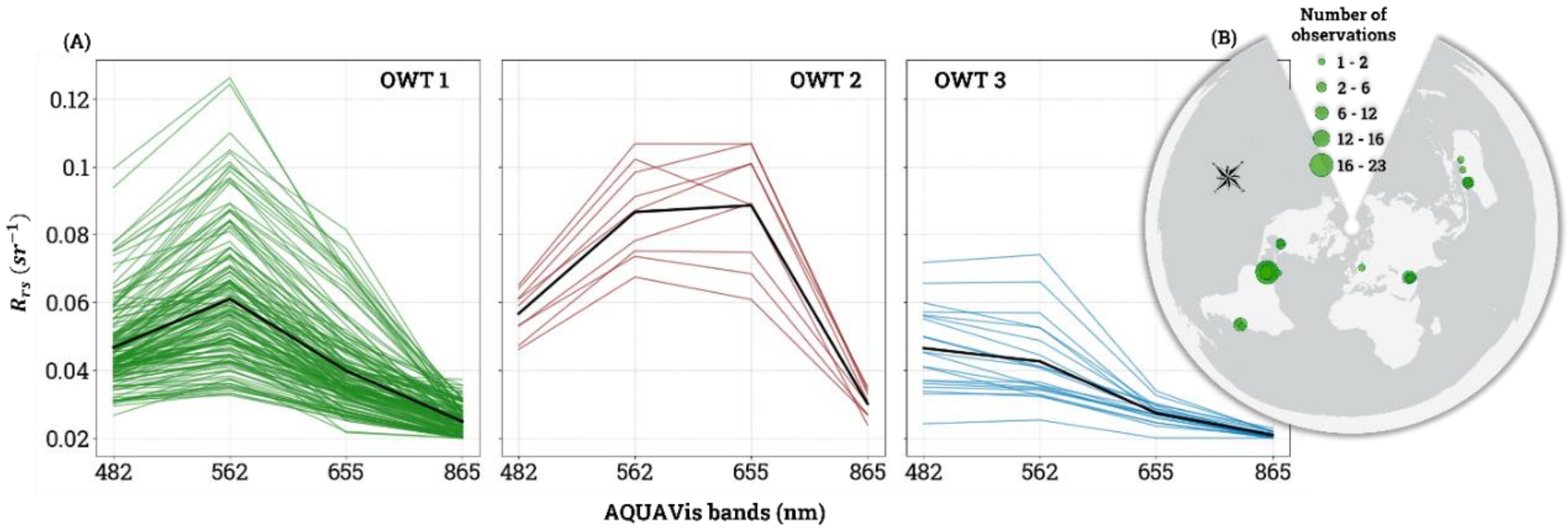

Figure S1 – (A) Mean Rrs spectra for each selected OLI/MSI image in the validation dataset, classified into the three defined Optical Water Types (OWTs). The black solid line represents the reference Rrs spectrum used in the Spectral Angle Mapper (SAM) classifier. (B) Geographical distribution of the selected images for validation and the number of algal bloom occurrences identified at each location.

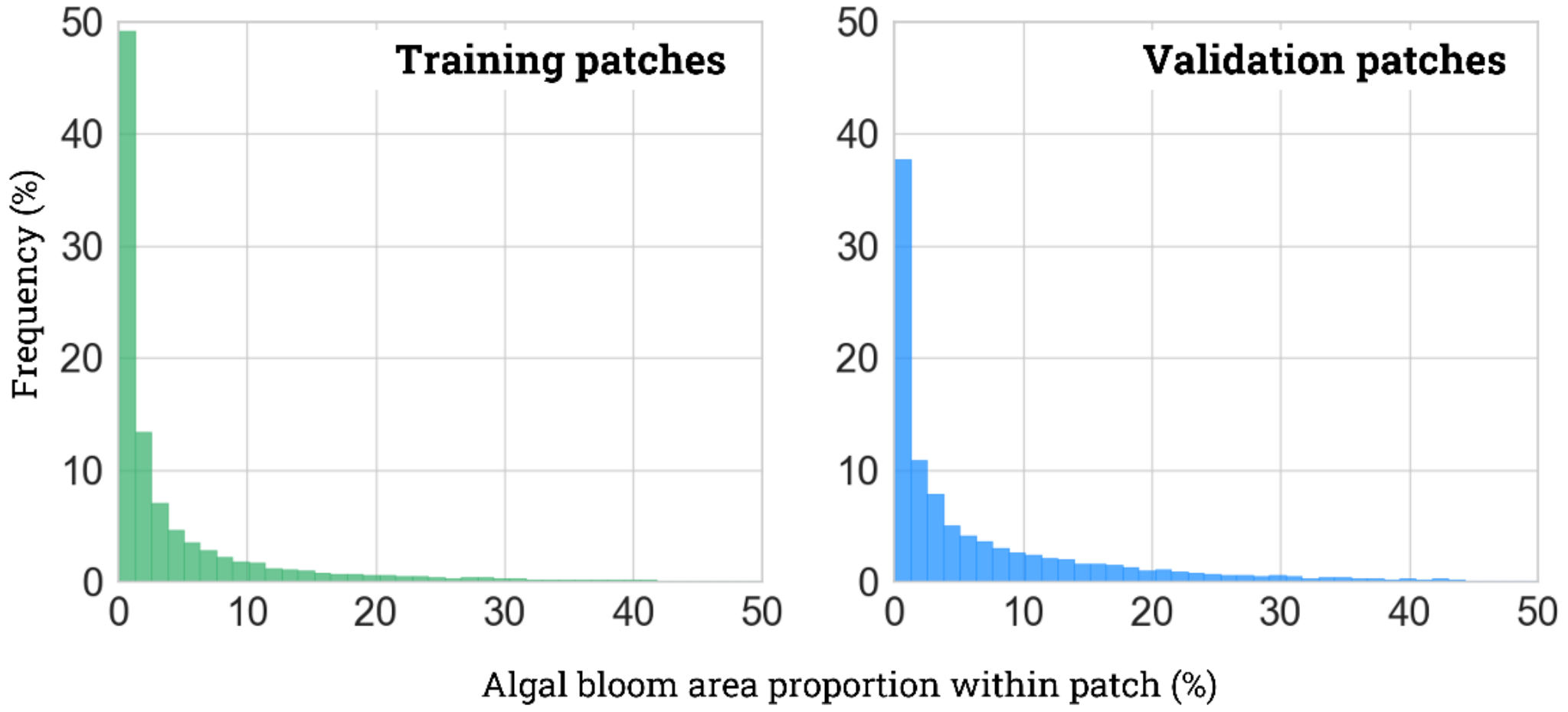

Figure S2 - Frequency distributions of the algal bloom area proportion (expressed as a percentage of the 256 × 256 30 m pixel patch) of the 24,265 training patches and the 12,255 validation patches.

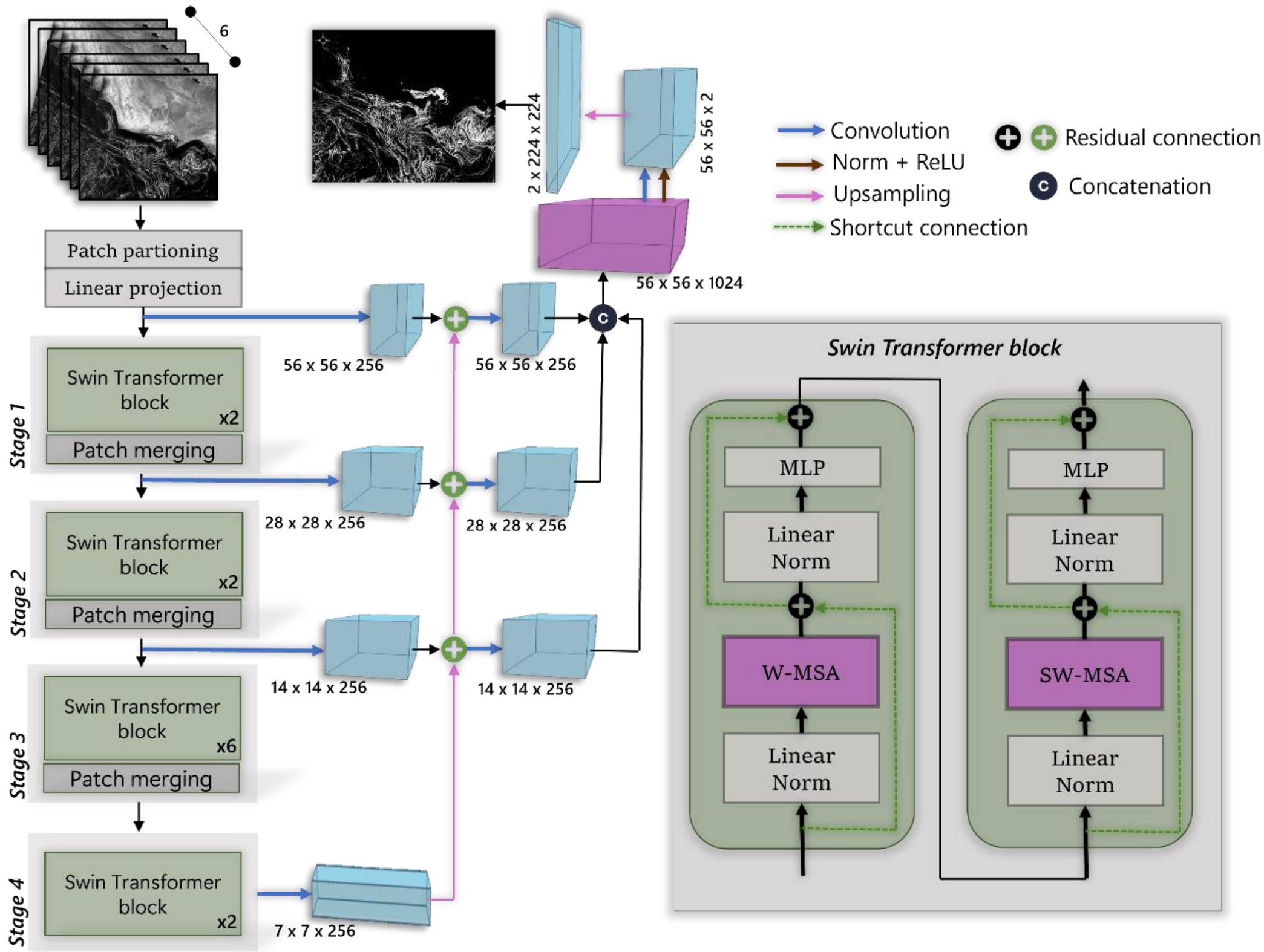

Figure S3. Architecture of the Swin Transformer model. The input image, resized to 224×224, is initially partitioned into non-overlapping 4×4 patches, each projected to a 96-dimensional token. The encoder comprises four hierarchical stages ([2, 2, 6, 2] ST blocks, green boxes), each integrating layer normalization, W-MSA, and MLP layers to produce four resolution outputs. Self-attention is computed locally within 7×7 windows using MHSA, with a subsequent half-window shift to capture global context (SW-MSA). Unlike Vanilla ViTs, spatial resolution decreases across stages through Patch Merging, which groups 2×2 adjacent patches, concatenates their embeddings, and applies a linear projection to increase channel dimensionality (Lin et al., 2022). The output of the ST backbone encoder (left side) consists of multi-layer feature maps from each stage, which were fused using a UPerNet-style decoder (Xiao et al., 2018; Zheng et al., 2021), beginning with a 1×1 convolution (blue arrows) to align channel dimensions across stages. Next, the decoder progressively upsample feature maps (pink arrows) from deeper layers and fuses them with higher-resolution features. Each fused feature map is also passed through a 3x3 convolution to refine local details and suppress artifacts introduced by upsampling. All scaled-adjusted feature maps are concatenated along the channel dimension and passed through a fusion head; a sequence of convolutional layers designed to aggregate the multi-scale context into a dense prediction map. The output is then upsampled to match the spatial resolution of the original input image, producing the final segmentation logits.

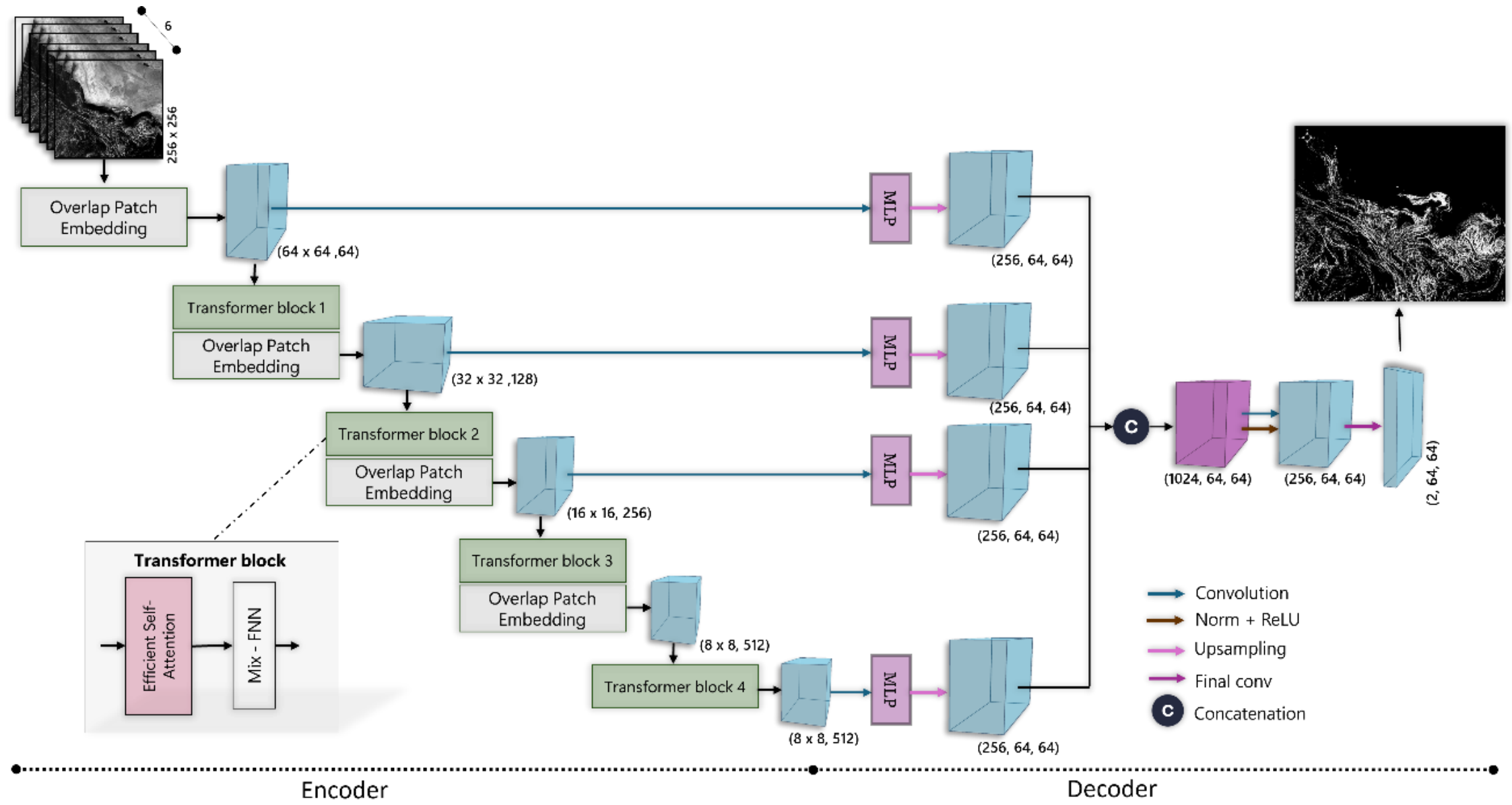

Figure S4. Architecture of the SegFormer model. Each stage begins with an overlapping patch embedding, allowing each patch to share information with its neighbors and capture local spatial context (Lin et al., 2023). The stages comprise a specific number of Transformer blocks (green blocks) with depths of [3, 4, 18, 3], where the two core components are the Efficient Self-Attention module, which downsamples the Key and Value tensors in equation (1) during attention calculation, and the Mix Feedforward Network (Mix-FFN), defined by a two-layer feed-forward CNN that injects localized spatial information into the attention outputs. Additionally, each Transformer block uses residual connections to improve training stability and generalization. The outputs of the encoder stages are passed to the decoder head, similar to the ST model, where all outputs are projected to a common dimension, enabling fusion across scales. Next, all feature maps are upsampled to match the first stage's output, concatenated along the channel dimension, and fused. A final 1×1 convolutional layer reduces the fused tensor to the desired number of output classes, producing dense per-pixel predictions. Finally, the model's output is upsampled using bilinear interpolation to match the original input resolution of 256×256.

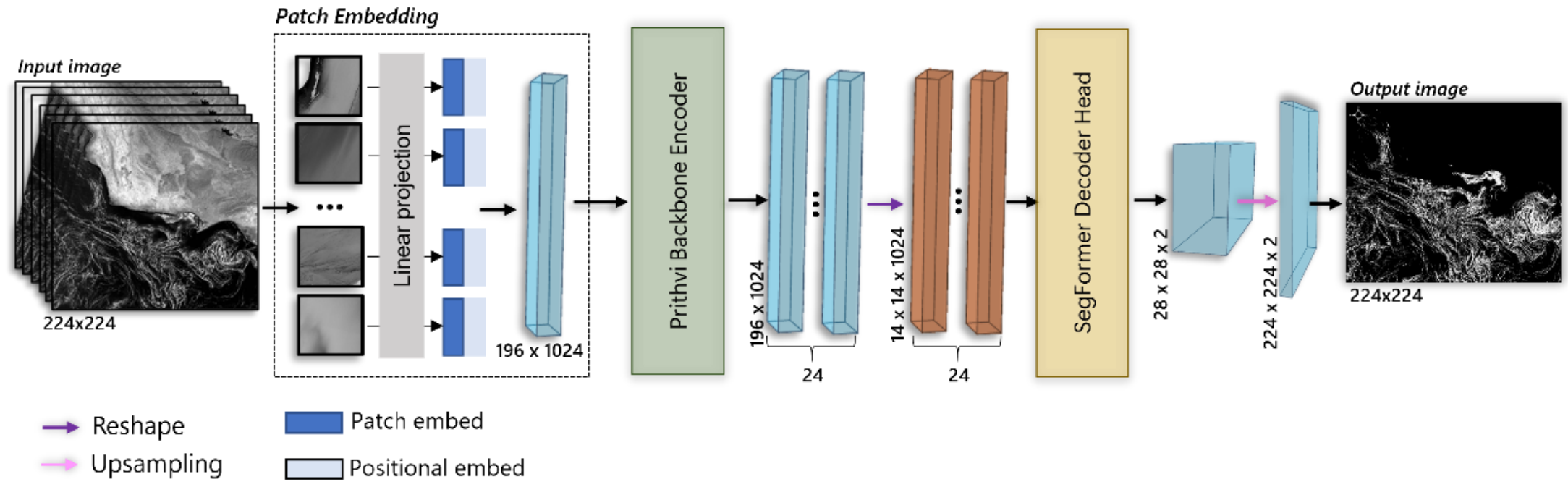

Figure S5. Overview of the Prithvi fine-tuning. The model receives an input image of size 224×224, which is split into non-overlapping patches. A linear projection maps each patch into a fixed-dimensional token embedding that are passed through a Transformer Encoder (green) consisting of 24 blocks. The encoder processes the tokens to produce latent feature representations, which are then passed to a SegFormer Decoder (yellow), which aims to reconstruct the original input.

Table S1. Evaluation metrics for algal bloom segmentation using five deep learning models: ResUNet, Vanilla ViT, Swin Transformer, SegFormer, and MAE (Prithvi). Metrics include recall, precision, F1-score (Dice coefficient), commission error (false positives), and omission error (false negatives). A buffer of 1 pixel was included in the ground mask, providing a confidence interval for models' prediction. Best performance values for each metric are highlighted in bold.

| Metrics | ResUNet | Vanilla ViT | Swin Transformer | SegFormer | MAE |
|---|---|---|---|---|---|
| **Recall** | 0.68 | 0.74 | 0.71 | **0.78** | 0.54 |
| **Precision** | 0.70 | 0.67 | **0.74** | 0.63 | 0.71 |
| **F1-Score (Dice)** | 0.69 | 0.70 | **0.73** | 0.70 | 0.62 |
| **Commission (%)** | 31.46 | 32.98 | **25.28** | 36.55 | 28.50 |
| **Omission (%)** | 29.57 | 26.17 | 28.39 | **21.91** | 45.45 |

Table S2. Sensitivity analysis of each DL model accuracy with respect to algal bloom area proportion. Results shown for six proportion groups composed of 12,255 256x256 30-meter patches per group. A buffer of 1 pixel was included in the ground mask, providing a confidence interval for models' predictions. Lowest errors per algal proportion are highlighted in bold.

| Error (%) | Model | Algal proportion (%) within patch | | | | | |
|---|---|---|---|---|---|---|---|
| | | **0 – 0.1** | **0.1 – 5** | **5 – 25** | **25 – 45** | **45 – 85** | **85 – 95** |
| **Omission** | **ResUNet** | 23.87 | 18.99 | 18.97 | 27.35 | 41.66 | 32.38 |
| | **Vanilla ViT** | **19.20** | 15.46 | 17.39 | 27.79 | 54.90 | 35.22 |
| | **Swin Transformer** | 26.87 | 16.49 | 17.22 | 25.12 | 42.20 | 25.48 |
| | **SegFormer** | 19.54 | **10.70** | **12.06** | **18.26** | **31.90** | **24.42** |
| | **MAE** | 28.53 | 28.20 | 36.06 | 53.08 | 77.79 | 60.09 |
| **Commission** | **ResUNet** | **3.01** | 16.88 | **25.69** | **32.35** | 41.06 | 41.67 |
| | **Vanilla ViT** | 5.13 | 22.70 | 34.30 | 43.40 | 48.23 | 47.72 |
| | **Swin Transformer** | 3.14 | **16.02** | 24.85 | 33.16 | 43.13 | 50.47 |
| | **SegFormer** | 4.37 | 22.42 | 33.79 | 40.88 | 46.42 | 48.11 |
| | **MAE** | 5.15 | 21.78 | 29.81 | 33.45 | **32.04** | **26.97** |

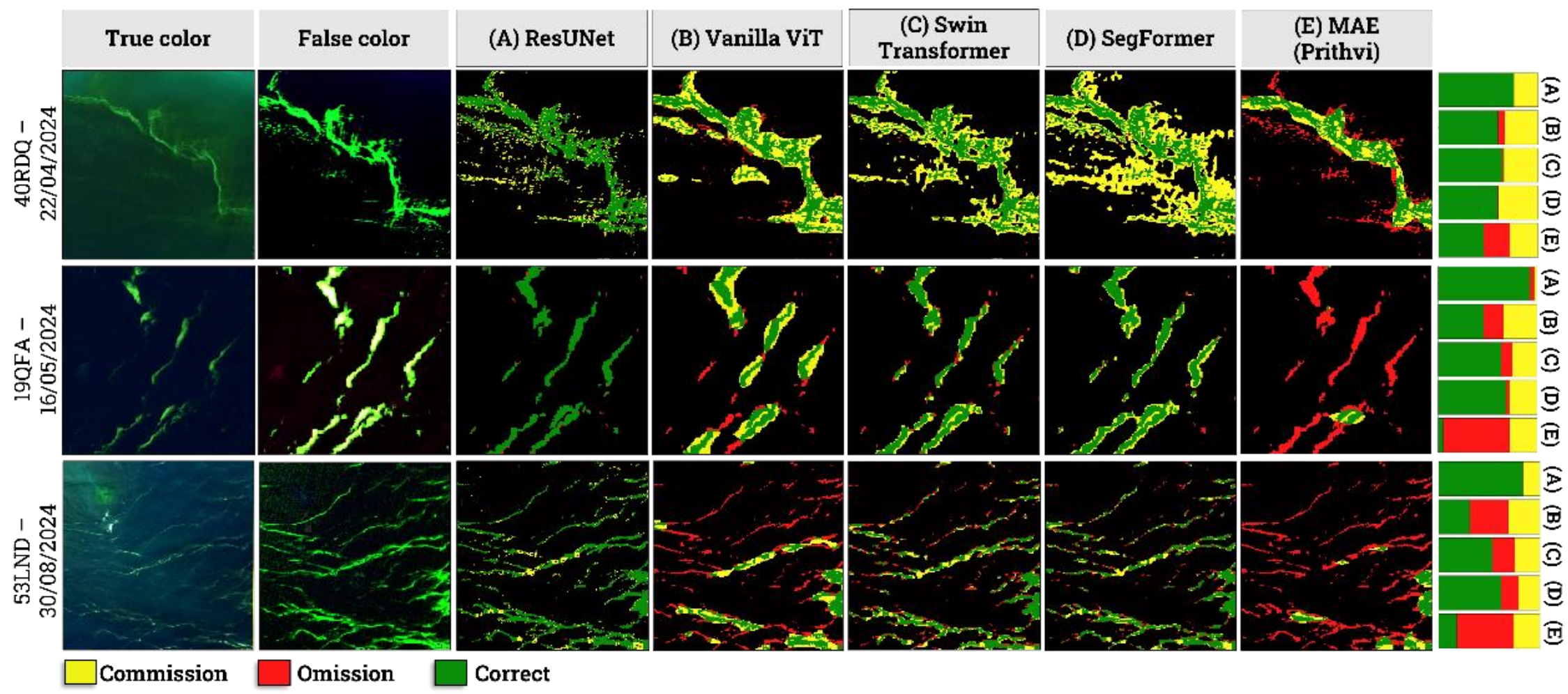

Figure S6. Examples illustrating segmentation results from each evaluated DL model. The first two columns *show* the true and false color compositions of the algal bloom occurrences, while the remaining columns *illustrate* the segmentation results and algal bloom reference masks together, showing correctly classified pixels (green), pixels that should have been classified as algal bloom but were not (red), and pixels that were incorrectly classified as blooms (yellow). The bar plots in the *rightmost* column *show* the correct, omission, and commission proportions for each model. The location of each example is indicated by its MGRS tile ID, with corresponding tile centroids as follows: 40RDQ (56.55°E, 26.62°N), 19QFA (67.53°W, 18.49°N), and 53LDN (135.51°E, 15.48°S).

Table S3. Summary of bloom records (2015–2022) cataloged by Gernez et al. (2023) and other authors, identified by Swin Transformer. For additional details regarding these records, please refer to Supplementary Material #2 in the original study.

| Class | Coastal area | Country | Date | Reference |
|---|---|---|---|---|
| Cyanophyceae | Albufera lagoon in Menorca | Spain | 20180709 | HAEDAT ES-18-014 |
| Cyanophyceae | Albufera lagoon in Menorca | Spain | 20180714 | HAEDAT ES-18-014 |
| Cyanophyceae | Albufera lagoon in Menorca | Spain | 20180724 | HAEDAT ES-18-014 |
| Cyanophyceae | Albufera lagoon in Menorca | Spain | 20180729 | HAEDAT ES-18-014 |
| Cyanophyceae | Jiquilisco Bay | El Salvador | 20190216 | HAEDAT SV-19-001 |
| Cyanophyceae | Baltic sea | Latvia | 20190720 | Zettler et al. (2020) |
| Cyanophyceae | Baltic sea | Sweden | 20190720 | Zettler et al. (2020) |
| Cyanophyceae | Baltic Sea | Sweden | 20150813 | HAEDAT SE-15-012 |
| Cyanophyceae | Baltic Sea | Sweden | 20190726 | HAEDAT SE-19-007, Karlson et al. (2021) |
| Cyanophyceae | Baltic Sea | Sweden | 20160725 | HAEDAT SE-16-006 |
| Pelagophyceae | Long Island Sound | US | 20160717 | HAEDAT US-16-009 |
| Dinophyceae | Chabahar Bay | Iran | 20171025 | Dolatabadi et al (2021) |
| Dinophyceae | Walker Bay | South Africa | 20190225 | Smith & Bernard (2020) |
| Dinophyceae | Walker Bay | South Africa | 20190225 | Smith & Bernard (2020) |
| Dinophyceae | Gulf of California | Mexico | 20180418 | HAEDAT MX-17-014 |
| Dinophyceae | West Coast of Florida | US | 20180728 | Weisberg et al. (2019) |
| Dinophyceae | West Coast of Florida | US | 20180807 | Weisberg et al. (2019) |
| Dinophyceae | Vilaine estuary | France | 20190706 | Roux et al. (2022) |
| Dinophyceae | Vilaine estuary | France | 20190716 | Roux et al. (2022) |
| Dinophyceae | La Jolla Bay | US | 20200415 | Kahru et al. (2021) |
| Dinophyceae | La Jolla Bay | US | 20200423 | Kahru et al. (2021) |
| Dinophyceae | La Jolla Bay | US | 20200425 | Kahru et al. (2021) |
| Dinophyceae | La Jolla Bay | US | 20200425 | Kahru et al. (2021) |
| Dinophyceae | La Jolla Bay | US | 20200425 | Kahru et al. (2021) |

| Dinophyceae | South Brittany | France | 20210814 | REPHY |
|---|---|---|---|---|
| Dinophyceae | Chesapeake Bay | US | 20200907 | HAEDAT US-20-016 |
| Litostomatea | Chabahar Bay | Iran | 20170408 | Ershadifar et al. (2020) |
| Litostomatea | Sables Olonne | France | 20200411 | Phenomer |
| Cyanophyceae | Laguna Lake | Philippines | 20200904 | Caballero & Navarro (2021) |
| Cyanophyceae | Laguna Lake | Philippines | 20200914 | Caballero & Navarro (2021) |
| Cyanophyceae | Laguna Lake | Philippines | 20200919 | Caballero & Navarro (2021) |
| Cyanophyceae | Lagoa dos Patos | Brazil | 20190227 | Qi et al. (2020) |
| Cyanophyceae | Lagoa dos Patos | Brazil | 20190301 | Qi et al. (2020) |
| Dinophyceae | NW Aegean Sea | Greece | 20190325 | HAEDAT GR-19-002 |
| Dinophyceae | Sables Olonne | France | 20200411 | Phenomer |
| Dinophyceae | Ria de Vigo | Spain | 20170925 | Blog fitopasion |
| Dinophyceae | Ria de Vigo | Spain | 20210904 | Blog fitopasion |
| Dinophyceae | Gran Canaria | Spain | 20160413 | Blog fitopasion |
| Dinophyceae | East China Sea | China | 20190608 | Caballero & Navarro (2021) |
| Dinophyceae | East China Sea | China | 20190608 | Qi et al. (2020) |
| Dinophyceae | East China Sea | China | 20190608 | Qi et al. (2020) |
| Dinophyceae | East China Sea | China | 20190608 | Qi et al. (2020) |
| Dinophyceae | Arabian Sea | Oman | 20190314 | Qi et al. (2020) |
| Dinophyceae | Arabian Sea | Oman | 20190314 | Qi et al. (2020) |
| Dinophyceae | Arabian Sea | Oman | 20190314 | Qi et al. (2020) |
| Cyanophyceae | Gulf of Gdańsk | Poland | 20180725 | HAEDAT PL-18-002 |
| Cyanophyceae | Gulf of Gdańsk | Poland | 20200626 | HAEDAT PL-20-001 |
| Cyanophyceae | Baltic Sea | Sweden | 20180731 | HAEDAT SE-18-005 |
| Cyanophyceae | Baltic Sea | Sweden | 20180731 | HAEDAT SE-18-005 |
| Cyanophyceae |  | Brazil | 20190201 | Qi et al. (2020) |
| Cyanophyceae | Guadeloupe | France | 20180328 | REPHY |
| Cyanophyceae |  | Australia | 20180603 | Gernez et al. (2023) |
| Cyanophyceae |  | Australia | 20180603 | Gernez et al. (2023) |
| Phaeophyceae | Caribean Sea | Haiti | 20220908 | Qi et al. (2025) |
| Ulvophyceae | Yellow Sea | China | 20190623 | Qi et al. (2020) |

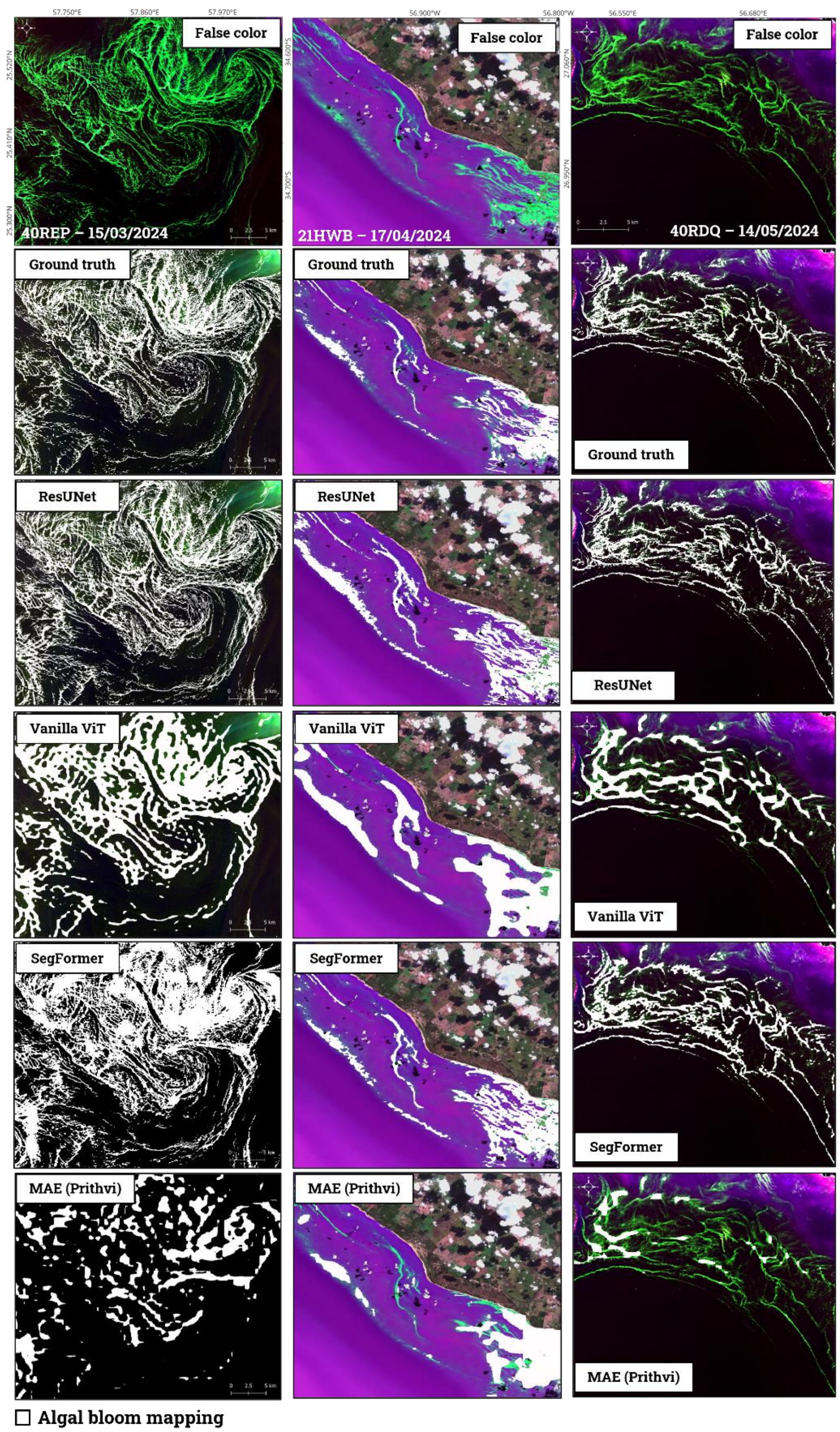

Figure S7. Segmentation examples for three OWTs using the CNN- and ViT-based models. For each OWT (see Figure S1), the false color composite (first row) shows the full scene, followed by the ground truth mask and the model's predictions. The location of each example is indicated by its MGRS tile ID, with corresponding tile centroids as

follows: 40REP (57.55°E, 25.72°N), 21HWB (56.40°W, 34.83°S), 40RDQ (56.54°E, 26.62°N).

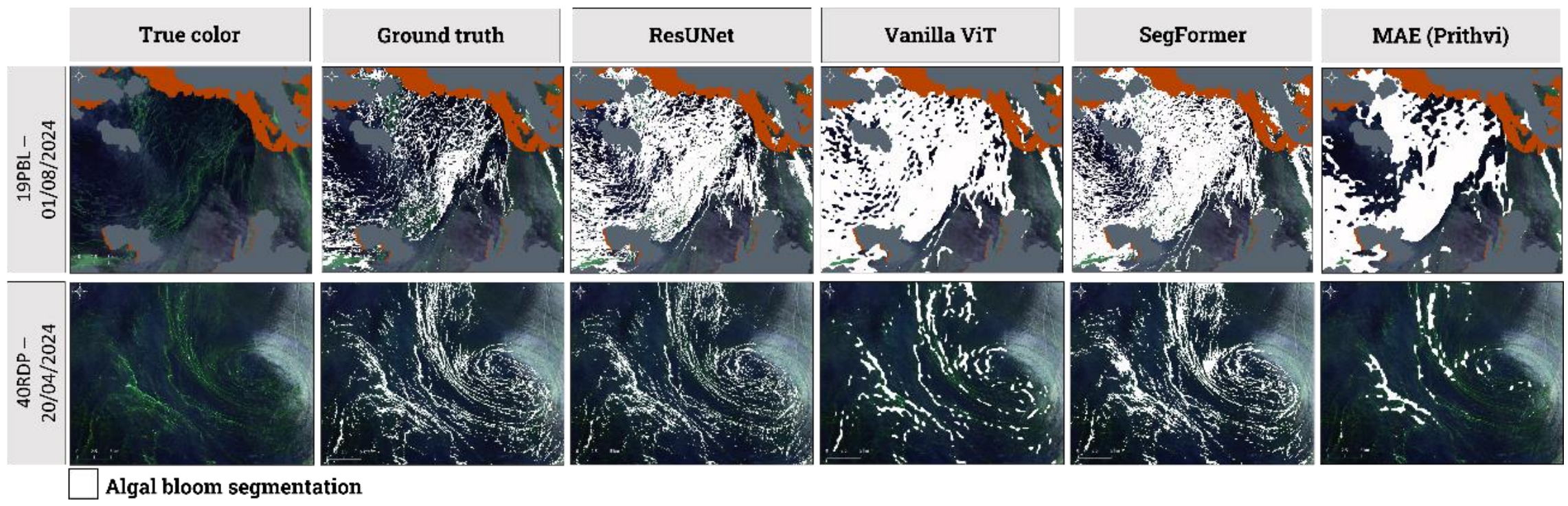

Figure S8. Segmentation examples under challenging atmospheric conditions. First row is a case dominated by high-altitude clouds, and second row is a case affected by strong glint. For each case the first column shows the true color composite, followed by the ground truth masks and the four CNN- and ViT-based models. Cloud and shadow masks are included in grey and orange, respectively. The location of each example is indicated by its MGRS tile ID, with corresponding tile centroids as follows: 19PBL (71.23°W, 9.44°N), 40RDP (56.55°E, 25.72°N).

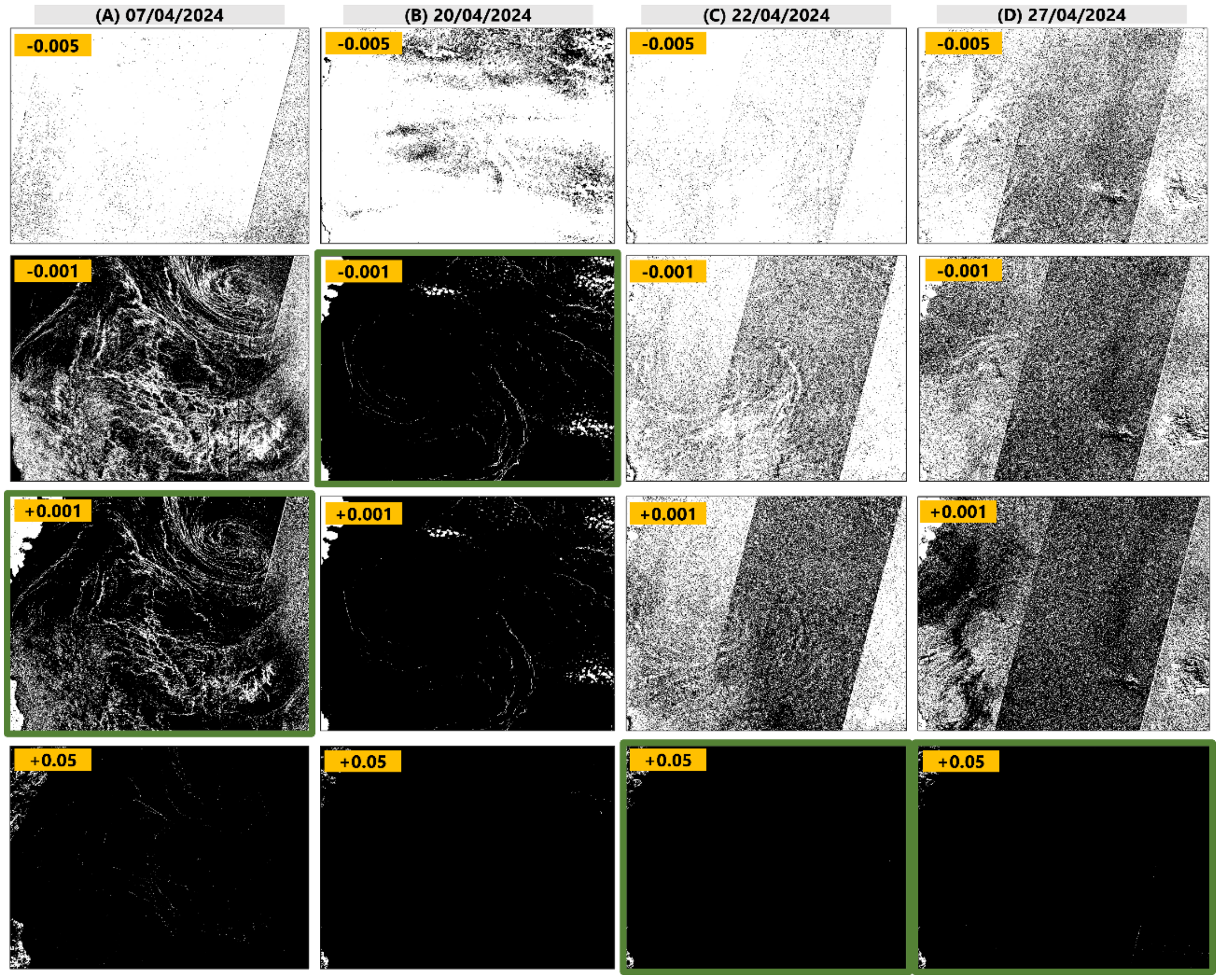

Figure S9. Sensitivity of FAI–based algal bloom detection using different thresholds across four acquisition dates. Each column corresponds to a single date, while each row shows the binary bloom mask (white = detected algae, black = non-bloom) generated using progressively increasing FAI thresholds (−0.005, −0.001, +0.001, and +0.05). The results illustrate substantial variability in optimal thresholding depending on scene

conditions, including atmospheric effects, water background variability, and sensor/viewing geometry. The thresholds considered most appropriate for each date are highlighted with green borders.

**Appendix S1: Pre-trained vanilla Vision Transformer**

Following the architecture illustrated in Figure 5, the input 256x256 patches were first resized to 224×224 pixels to match the pre-trained weights, and partitioned into non-overlapping sub-patches of size 16×16, yielding a total of 196 sub-patches per patch. Each patch was flattened and linearly projected into a 384-dimensional embedding space through a trainable projection layer. This results in a tensor $X$ of shape [B, 196, 384], where B is the batch size. Positional embeddings were added to the tokens to preserve spatial relationships, and the resulting sequence was processed through a series of 8 transformer encoder blocks (Dosovitskiy et al., 2021). Each block comprises 6 multi-head self-attention (MHSA) and Multilayer Perceptron (MLP) sub-layers. MHSA allows the model to learn information from different representation subspaces by computing different attention weights across all token pairs in parallel. Within each MHSA module, the tensor $X$ is linearly projected into three new vectors called queries (Q), keys (K), and values (V) (Vaswani et al., 2017):

$$Q = XW^{Q}, K = XW^{K}, V = XW^{V} \tag{1}$$

Where $X \in R^{128\times196\times384}$ as previously explained, and $W^{Q,K,V}$ are learned projection matrices. Conceptually, $Q$ encodes what each token is seeking in another token, while $K$ represents what each token offers, and $V$ contains the actual semantic content of the tokens (Niu et al., 2021). The attention mechanism computes pairwise interactions by taking the dot product between each query and key, producing similarity scores that are scaled and passed through a softmax function to obtain attention weights (**Error! Reference source not found.**, right panel). These weights are then used to compute a weighted sum of the value vectors, allowing the model to aggregate context-aware information from relevant tokens (Vaswani et al., 2017):

$$Attention(Q, K, V) = softmax\left(\frac{QK^{T}}{\sqrt{d_k}}\right)V \tag{2}$$

$d$ is the dimension of (query, key, value). The attention outputs of each head are concatenated and linearly transformed to form the final attention output. At each epoch, the attention mechanism updates token representations based on pairwise relevance, which are then refined by MLP layers to enhance spatial understanding (Figure 5, left side) (Dosovitskiy et al., 2021). While the transformer encoder preserves the token grid size, the initial patch embedding downscales the input. The decoder then progressively upsamples the latent representations to recover the original resolution, using projected and interpolated skip connections from selected transformer layers for multi-scale feature fusion and spatial detail preservation (Figure 5, right side).

**Appendix S2: Impact of Spectral Index Combinations on Model Performance**

To assess the role of spectral indices in model performance, a comparative experiment was conducted with Swin Transformer using three input configurations: VNIR+FAI+NDVI, VNIR+NDVI, and VNIR-only. All models were trained using identical hyperparameters and procedures to ensure consistency (Section 3.4.3). Performance metrics for the 2024 validation dataset are provided in Table S4, while Figure S10 provides visual comparisons of the predicted bloom masks.

Table S4. Evaluation metrics for algal bloom segmentation using Swin Transformer trained on three input channel configurations. Metrics include recall, precision, F1-score (Dice coefficient), commission error (false positives), and omission error (false negatives).

| Metrics | VNIR + NDVI + FAI | VNIR + NDVI | VNIR |
|---|---|---|---|
| **Recall** | 0.87 | 0.87 | 0.89 |
| **Precision** | 0.42 | 0.41 | 0.37 |
| **F1-Score (Dice)** | 0.57 | 0.56 | 0.52 |
| **Commission (%)** | 57.51 | 59.26 | 62.94 |
| **Omission (%)** | 13.12 | 12.34 | 10.20 |

From Table S4, the inclusion of NDVI and FAI leads to only marginal changes in overall accuracy. However, such metrics do not fully reflect their functional role. As shown in Fig. S10, these additional channels primarily affect the spatial representation of the predictions, leading to improved delineation of filament structures and reduced commission errors along bloom boundaries. Traditional indices are typically applied to such mappings using fixed thresholds on individual pixels, whereas deep learning models rely on spatial relationships of features represented in the image (Blondeau-Patissier et al., 2014; Ronneberger et al., 2015). In transformer-based architectures, for example, self-attention weights are calculated by relative token-to-token contrast rather than absolute radiometric magnitude (Vaswani et al., 2017; Dosovitskiy et al., 2021). This shift from absolute values to relative context fundamentally redefines the role of spectral information during training: the model learns to identify features based on their contrast against the background, and thus it becomes inherently resilient to scene-wide additive offsets. In optically complex waters, uncertainties in the atmospheric correction or residual sun glint effects mainly introduce additive offsets in NDVI or FAI values, but they can still preserve the spatial contrast and inherent structure of blooms. Consequently, the input $R_{rs}$ bands and the spectral indices serve as high-contrast channels that remain informative despite uncertainties in absolute magnitude (Hu et al., 2019; Wang and Hu, 2021). This potentially explains why NDVI and FAI, although not significantly improving accuracy metrics, consistently lead to more consistent bloom delineation in optically complex waters.

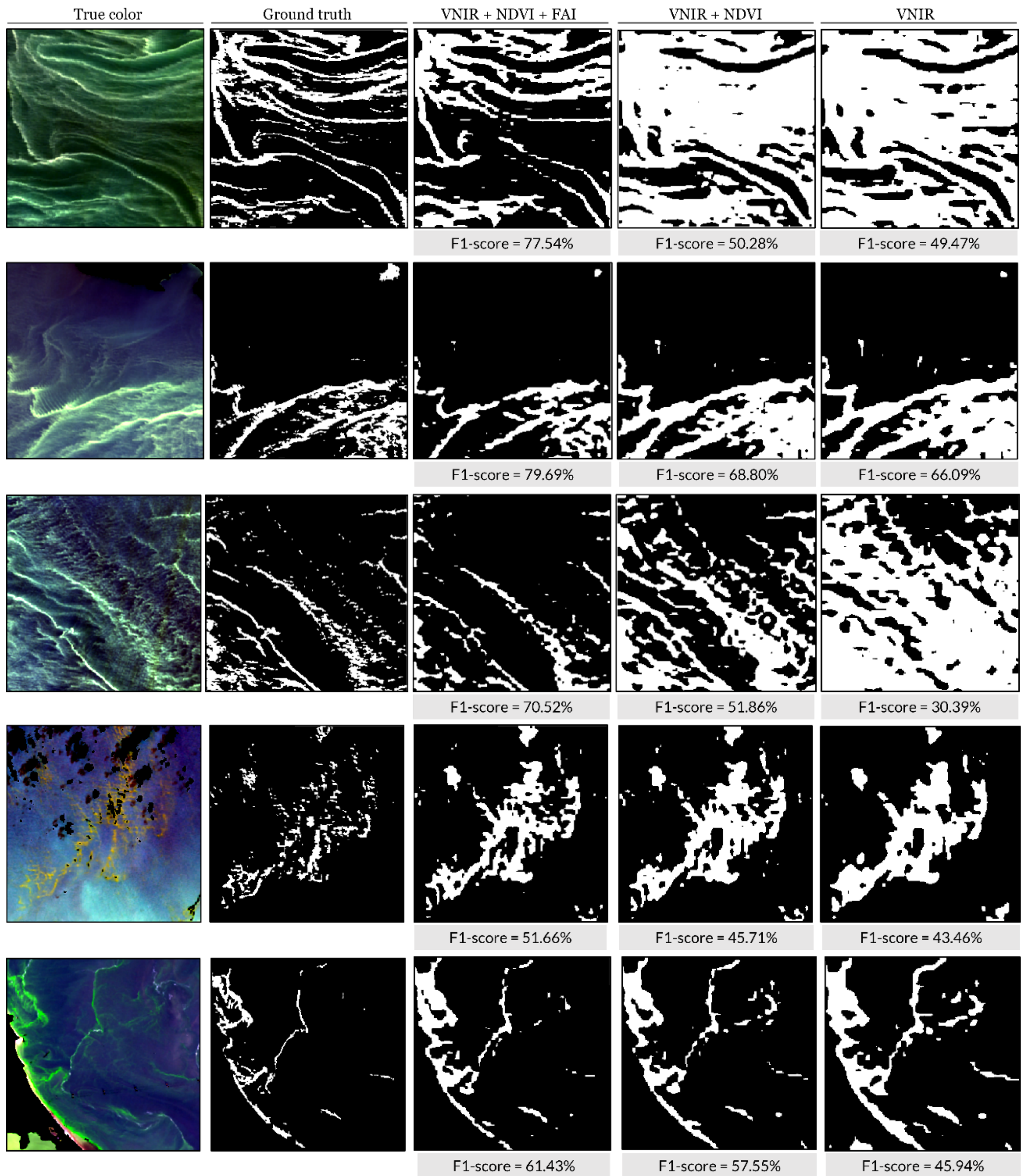

Figure S10. Qualitative analysis comparing algal bloom mask predictions from three different sets of input channels for training: (1) VNIR+NDVI+FAI, (2) VNIR+NDVI, and (3) VNIR-only. The first column corresponds to true color composition, followed by the ground truth mask and each evaluated model. For each model, patch-based F1-scores are shown at the bottom of the segmentation masks.

## Appendix S3: Assessing the impact of radiometric uncertainty on final model predictions

Remote sensing reflectance is inherently susceptible to several uncertainties, such as atmospheric correction, adjacency effects, and sun glint, and can potentially affect the algal bloom mapping quality. To examine how such uncertainties influence model performance, we evaluated the impact of noisy input data on prediction robustness, following the approach of Yuan et al. (2025). Here, robustness refers to the ability of a model to maintain consistent performance under both nominal and perturbed conditions, which is critical for practical applications. A standard method for assessing this resilience is the evaluation of performance on data stressed with Gaussian noise (Grandvalet et al., 1997; Narayanan et al., 2003; Sadeghzadeh et al., 2026). Here, we utilized the bias-corrected Root Mean Square Error (RMSE) per band from AQUAVis products ($\Delta R_{rs_{blue}} = 0.0089$; $\Delta R_{rs_{blue}} = 0.0078$; $\Delta R_{rs_{blue}} = 0.0058$; $\Delta R_{rs_{blue}} = 0.0022$) to simulate realistic sensor noise. The noisy inputs were generated as follows:

$$R_{rs,noisy} = R_{rs,original} + \varepsilon, \quad \varepsilon \sim N(0, \sigma^2)$$

Where $\varepsilon$ represents the Gaussian noise with a mean of zero and variance $\sigma^2$. The injected noise was proportional to the standard deviation of each input at three levels: 10%, 20%, and 30%. The 20% level aligns with typical retrieval uncertainties, particularly in cloudy or heterogeneous regions, while the 10% and 30% levels provided a sensitivity range to illustrate performance variation under lower and higher uncertainty conditions. To ensure statistical significance, we applied 50 Monte Carlo simulations per patch using the validation dataset and aggregated the resulting prediction outputs to compare with the baseline input image.

Table S4. Performance of the Swin Transformer model under increasing levels of injected Gaussian noise, calibrated to per-band uncertainty of the AQUAVis products. Metrics are reported for the baseline (no noise) and three noise levels (10%, 20%, and 30% of each band's standard deviation), aggregated over 50 Monte Carlo samples per patch on the validation dataset.

| Metrics | Baseline | 10% | 20% | 30% |
|---|---|---|---|---|
| **Recall** | 0.87 | 0.85 | 0.81 | 0.76 |
| **Precision** | 0.42 | 0.44 | 0.47 | 0.49 |
| **F1-Score (Dice)** | 0.57 | 0.58 | 0.59 | 0.60 |
| **Commission (%)** | 57.51 | 55.79 | 53.03 | 51.02 |
| **Omission (%)** | 13.12 | 14.77 | 18.75 | 23.32 |

Results in Table S4 demonstrate that overall model performance remained stable across all levels of noise perturbations, with F1-score holding near constant from 0.57 at baseline to 0.60 at 30% of noise level. At the 20% noise level, commission and omission errors showed mean differences from the baseline of only -3.8% and 3.3%, respectively (Figure S11A). Spatially, Figure S11B corroborates this pattern, where the difference maps reveal that at 10% noise, deviations are minimal and spatially diffuse, suggesting that low-level uncertainty has a negligible impact on the predicted segmentation structure. At the 30% noise level, performance shifts are concentrated primarily along the high-frequency edges of algae patch boundaries. As also observed on the additional examples in Figure S12, these deviations do not exhibit the structured spatial patterning characteristic of systematic artifacts, such as scan-line effects. This indicates that the observed variations reflect a genuine sensitivity to input uncertainty rather than the propagation of model-induced artifacts, confirming the Swin Transformer's robustness to realistic radiometric noise.

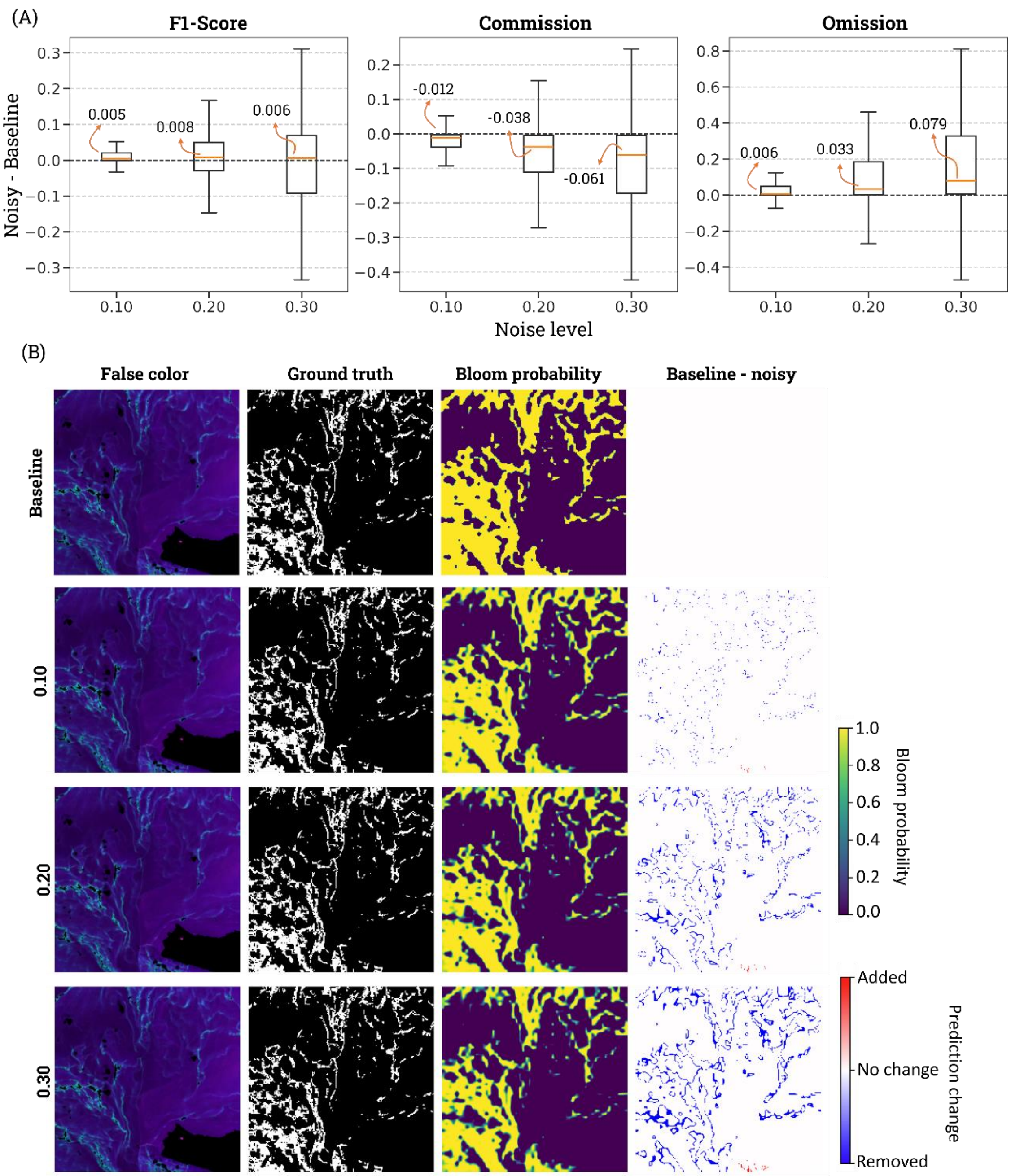

Figure S11. Qualitative assessment of Swin Transformer robustness to radiometric uncertainty. (A) Boxplot of differences between baseline metrics (no noise) and noisy metrics (10%, 20%, and 30%). (B) Patch example illustrating models' performance for each noise level. Each row corresponds to a noise level: baseline (no noise), 10%, 20%, and 30% of the per-band standard deviation derived from AQUAVis retrieval uncertainty. Columns show, from left to right: the false-color input image, the reference label mask, the model prediction probability, and the spatial difference map between the noisy prediction and the baseline prediction. Difference maps highlight pixels where noise-induced perturbations altered the segmentation outcome, with blue indicating pixels lost relative to baseline (omission shifts) and red indicating pixels gained (commission shifts).

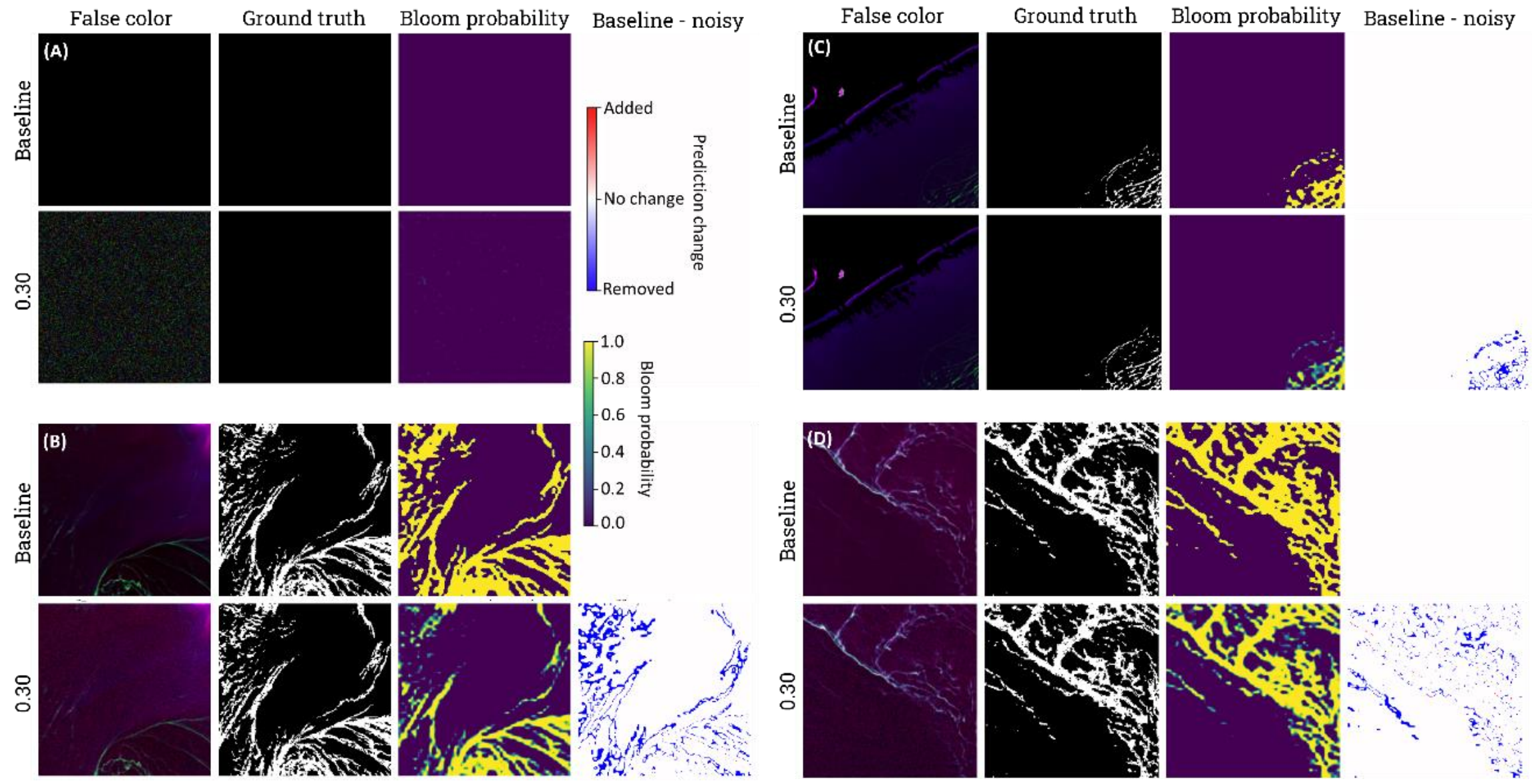

Figure S12. Patch example illustrating models' performance for 30% of noise level. Each row corresponds to a noise level: baseline (no noise) and 30% of the per-band standard deviation derived from AQUAVis retrieval uncertainty. Columns show, from left to right: the false-color input image, the reference label mask, the model prediction probability, and the spatial difference map between the noisy prediction and the baseline prediction. Difference maps highlight pixels where noise-induced perturbations altered the segmentation outcome, with blue indicating pixels lost relative to baseline (omission shifts) and red indicating pixels gained (commission shifts).